%% file: paper.tex
\documentclass{article}
\usepackage{arxiv,times}

\input{math_commands.tex}

\usepackage{hyperref}
\usepackage{url}

\usepackage{booktabs}
\usepackage{xspace}
\usepackage{amsmath}
\usepackage{amssymb}
\usepackage{siunitx}
\usepackage{cleveref}
\crefname{figure}{Figure}{Figures}
\crefname{table}{Table}{Tables}
\crefname{equation}{Eq.}{Eqs.}
\crefname{section}{Section}{Sections}
\crefname{appsec}{Appendix}{Appendices}
\crefname{appendix}{Appendix}{Appendices}
\crefname{subappendix}{Appendix}{Appendices}
\crefname{subsubappendix}{Appendix}{Appendices}

\usepackage{array}
\usepackage{enumitem}
\usepackage{graphicx}
\usepackage[version=4]{mhchem}
\usepackage{multirow}
\usepackage{wrapfig}
\usepackage[table]{xcolor}

\newcommand{\dataname}{FREA\xspace}
\newcommand{\discname}{BARTSmiles (disc.)\xspace}
\newcommand{\negname}{BARTSmiles (LL $+$ neg.)\xspace}
\newcommand{\gbdtname}{Descriptor GBDT\xspace}
\newcommand{\aizname}{AiZynthFinder filter\xspace}

\newcommand{\negrrfull}{Random Replacement\xspace}
\newcommand{\negrrabbr}{RR\xspace}
\newcommand{\negftfull}{Forward Template\xspace}
\newcommand{\negftabbr}{FT\xspace}
\newcommand{\negrafull}{Random Addition\xspace}
\newcommand{\negraabbr}{RA\xspace}
\newcommand{\negncfull}{New Center\xspace}
\newcommand{\negncabbr}{NC\xspace}
\newcommand{\negetfull}{Elementary Template\xspace}
\newcommand{\negetabbr}{ET\xspace}
\newcommand{\rulemethodname}{Rule Check\xspace}

\definecolor{groupheader}{gray}{0.9}

\title{\dataname: A Multi-Source Expert Benchmark for\\Reaction Feasibility Verification}

\author{Botao Yu$^{1,\dagger}$, Bo Zhou$^{2,*}$, Daniel Adu-Ampratwum$^{3,*}$, Frazier N. Baker$^{1,4}$, Ziru Chen$^{1}$, \\
\textbf{Reza Averly$^{1}$, Ye Liu$^{4}$, Wenhao Gao$^{5}$, Xia Ning$^{1,3,4,6}$, Huan Sun$^{1,\dagger}$} \\[0.3em]
$^{1}$Department of Computer Science and Engineering, The Ohio State University \\
$^{2}$Department of Pharmaceutical Sciences, University of Illinois Chicago \\
$^{3}$College of Pharmacy, The Ohio State University \\
$^{4}$Department of Biomedical Informatics, The Ohio State University \\
$^{5}$Department of Chemical and Biomolecular Engineering, University of Pennsylvania \\
$^{6}$Translational Data Analytics Institute, The Ohio State University
}

\newcommand{\goldN}{\num{751}}
\newcommand{\goldPos}{\num{341}}
\newcommand{\goldNeg}{\num{410}}
\newcommand{\goldAgreePct}{77.9\%}
\newcommand{\goldKappa}{0.55}
\newcommand{\goldKappaCI}{[0.49, 0.61]}
\newcommand{\goldKappaN}{\num{759}}
\newcommand{\goldPoolN}{\num{800}}
\newcommand{\goldCalibInPool}{\num{41}}

\begin{document}

\maketitle
\def\thefootnote{*}\footnotetext{Equal contribution.}
\def\thefootnote{\dag}\footnotetext{Correspondence to: \texttt{\{yu.3737, sun.397\}@osu.edu}.}
\def\thefootnote{\arabic{footnote}}

\begin{abstract}
As generative models and AI agents propose chemical reactions at a scale beyond expert review, feasibility verifiers decide which proposals enter synthesis planning. But do their decisions agree with chemists across different kinds of candidates?
We introduce \dataname, a benchmark of \goldN{} reactions labeled by expert chemists under an explicit feasibility criterion, drawn from retrosynthesis model proposals, zero-yield experimental records, edits by large language models (LLMs), and five negative candidate generation methods.
Our evaluation finds that no verifier leads across all sources: LLMs given only the criterion are competitive with dedicated verifiers, while forward models perform best on retrosynthesis proposals but reject most feasible edits of recorded reactions at the evaluated operating points.
Looking beyond aggregate scores, both forward models perform below chance when separating infeasible alternative disconnections from feasible generated candidates.
To study whether negative supervision addresses these weaknesses, we also release a corpus of over 14 million recorded reactions and generated negative candidates. In matched training comparisons, adding a mixture of generated negatives to forward training raises mean AUROC across sources, but these gains do not extend to retrosynthesis proposals. Varying the generation method further shows that the largest gain on generated candidates coincides with worse proposal screening.
These findings motivate evaluating verifiers against experts across sources and designing negatives for transfer to model proposals.
The benchmark and corpus are publicly available.
\end{abstract}

\input{sections/intro.tex}

\input{sections/related_work.tex}

\input{sections/task.tex}

\input{sections/dataset.tex}

\input{sections/protocol.tex}

\input{sections/results.tex}

\input{sections/negatives.tex}

\input{sections/conclusion.tex}

\input{sections/acknowledgments.tex}

\bibliography{reference}
\bibliographystyle{arxiv}

\clearpage
\appendix
\counterwithin{table}{section}
\counterwithin{figure}{section}
\renewcommand{\theHtable}{\thesection.\arabic{table}}
\renewcommand{\theHfigure}{\thesection.\arabic{figure}}

\input{appendices/related_work.tex}

\input{appendices/annotation.tex}

\input{appendices/data_curation.tex}

\input{appendices/negative_generation_details.tex}

\input{appendices/system_implementations.tex}

\input{appendices/protocol_details.tex}

\input{appendices/additional_results.tex}

\input{appendices/data_card.tex}

\input{appendices/limitations.tex}

\end{document}

%% file: math_commands.tex
\usepackage{amsmath,amsfonts,bm}

\def\eqref#1{equation~\ref{#1}}

\def\1{\bm{1}}

\DeclareMathAlphabet{\mathsfit}{\encodingdefault}{\sfdefault}{m}{sl}
\SetMathAlphabet{\mathsfit}{bold}{\encodingdefault}{\sfdefault}{bx}{n}

%% file: sections/intro.tex
\section{Introduction}
\label{sec:intro}

Generative models, search-based planning systems and language-model agents now propose
chemical reactions at a scale that no chemist can review one by one
\citep{segler20183nmcts,long2025retrosynthesis,baker2025larc}. Within these pipelines,
reaction feasibility verifiers act as gatekeepers that decide which proposals survive
into downstream planning \citep{segler20183nmcts,genheden2020filter,sadowski2025retrotrim}.
Accepting an infeasible reaction can derail a plan, while rejecting a feasible
alternative can eliminate a useful route. Reliable verification therefore becomes
increasingly important as human review is removed from the loop.

Common verifiers include forward models, which use target-product recovery or likelihood
as a proxy for feasibility \citep{schwaller2019evaluation,schwaller2020predicting},
and classifiers trained to separate recorded reactions from generated negatives
\citep{segler20183nmcts,genheden2020filter,gainski2025retrogfn}, which are typically
assessed by their effect on downstream search or on held-out generated reactions
(\cref{app:related}).
Here, we ask: \textit{do verifiers that appear reliable on
recorded or generated reactions agree with chemists on the reactions they are actually
asked to screen}?

Answering this question requires labels that are independent of how verifiers are built.
Published reaction data favor successful outcomes \citep{voinarovska2023yield}, while
automatically generated negative candidates need not be infeasible. Expert-labeled
reactions therefore provide a direct basis for evaluating feasibility verifiers.

We introduce \dataname, an expert-labeled benchmark of \goldN{} reactions. A
reaction is feasible if it lists all substantive reactants and admits reasonable
conditions that yield the target products in isolable form (\cref{sec:task}). The reactions
come from four sources: proposals of eight retrosynthesis models, zero-yield experimental
records, edits of recorded reactions by large language models (LLMs), and candidates from
five negative candidate generation methods (\cref{sec:benchmark}). Annotators label each
reaction without seeing its source.

Comparing LLMs, forward models and verifiers trained on generated negatives, we find that
\textbf{no verifier leads across all sources} (\cref{subsec:overall}). LLMs given only
the criterion are competitive with dedicated verifiers, with Claude Opus 5 achieving the
highest mean balanced accuracy. Forward models are strongest on retrosynthesis proposals,
yet at the evaluated operating points they reject most feasible edits of recorded reactions.

Looking within the generated source reveals a sharper failure: \textbf{the strongest
verifiers of model proposals can rank infeasible candidates above feasible ones}.
Both forward models score below chance on alternative product disconnections, with
AUROCs of 26.2 and 32.5 against the same feasible reference reactions
(\cref{subsec:per_method}). Adding generated negatives to forward training improves this
comparison and gives the highest mean AUROC across sources, but reduces performance on
other constructions. These selective gains leave open whether negative supervision
improves verification beyond particular constructions, especially on model proposals.

To test this, we use an accompanying corpus of over 14 million recorded reactions and
generated negative candidates, with no benchmark reaction in the training data
(\cref{app:splits}). In matched comparisons with continued forward training,
\textbf{the mean AUROC gain from a mixture of generated negatives does not extend to
retrosynthesis proposals}
(\cref{sec:negatives}). Varying the generation method further shows that the largest gain
on generated candidates coincides with a decline on model proposals.
The open challenge is therefore to generate negatives whose benefits extend to the
proposals verifiers must screen.

Our contributions are:
\begin{enumerate}[leftmargin=1.5em, itemsep=2pt, topsep=3pt]
  \item \textbf{An expert benchmark} of \goldPos{} feasible and \goldNeg{} infeasible
    reactions from four sources, labeled under an explicit criterion and released with
    recorded failure reasons (\cref{sec:task,sec:benchmark}).
  \item \textbf{An evaluation of verifier strengths and failure modes}, across reaction
    sources and candidate generation methods
    (\cref{sec:protocol,sec:results}).
  \item \textbf{A generated corpus and a controlled transfer study}, with matched and
    pre-registered training comparisons that test whether negative supervision improves
    screening of model proposals and motivate hypotheses for negative-data design
    (\cref{sec:negatives}).
\end{enumerate}
The benchmark\footnote{\url{https://huggingface.co/datasets/FREA-benchmark/FREA-benchmark}}
and generated corpus\footnote{\url{https://huggingface.co/datasets/FREA-benchmark/FREA-corpus}}
are available on Hugging Face.

%% file: sections/related_work.tex
\section{Related Work}
\label{sec:related}

\paragraph{Feasibility filters.}
Classifiers trained on recorded reactions and constructed negatives have been
used in synthesis planning since the in-scope filter of
\citet{segler20183nmcts}. AiZynthFinder uses template-generated
negatives~\citep{genheden2020aizynthfinder,genheden2020filter}.
RetroGFN combines forward-template negatives with replacements of the
recorded product~\citep{gainski2025retrogfn}.
RetroTrim combines several scorers and evaluates both individual reactions
and complete routes by expert review~\citep{sadowski2025retrotrim}.
These studies establish the value of filtering in planning, but evaluation on
generated negatives does not test whether the assigned negative labels agree
with experts.
\dataname evaluates individual reactions across four sources under a common
expert criterion.

\paragraph{Expert and precedent-based evaluation.}
ChemCensor scores reactions by checking their reaction centers and
functional-group context against recorded reactions
\citep{zagribelnyy2026chemcensor}. Concurrent work also evaluates retrosynthesis
predictions against expert judgments~\citep{zagribelnyy2026ursa}.
\dataname combines model proposals with experimental records and multiple
forms of controlled perturbation under an explicit feasibility criterion.
This broader coverage supports both diagnosis of source-dependent failures
and evaluation of transfer from generated negative training data.

\paragraph{Negative reaction data.}
Published failures are scarce~\citep{voinarovska2023yield}. Existing
resources include private laboratory records~\citep{jiang2021smiles},
high-throughput experiments for acid--amine
couplings~\citep{zhong2025feasibilityHTE}, and data for enzymatic
reactions~\citep{chainani2025dora}. Experimental failures describe particular
conditions. \dataname uses zero-yield records as candidates for expert
annotation and combines them with model proposals and constructed reactions.
\Cref{app:related} gives a detailed comparison of these resources and
adjacent approaches.

%% file: sections/task.tex
\section{Reaction Feasibility Task Definition}
\label{sec:task}

Given reactants $\mathbf{R} = \{r_1,\ldots,r_k\}$ and products
$\mathbf{P} = \{p_1,\ldots,p_m\}$, a verifier predicts
\begin{equation}
    f(\mathbf{R},\mathbf{P}) \in \{0,1\},
\end{equation}
where 1 denotes a feasible reaction and 0 an infeasible one. The input contains no
reaction conditions, and $\mathbf{P}$ need not include every product of the reaction.

We define feasibility from the perspective of synthesis planning, in which a verifier
screens retrosynthesis proposals. Each reaction is treated as a proposal, which is feasible
if a planning system could keep it as a valid step of a route.
Specifically, a feasible proposal must satisfy two conditions.

\paragraph{Completeness.}
Completeness asks whether the proposal lists everything the step needs, apart from
routine components. The proposal must list every substantive reactant required to
produce the target products, rather than only a transformation from a main reactant to
a product. Auxiliary components, which act in routine supporting roles such as
catalysts, solvents, bases, reducing agents and water, may be assumed. Any species
outside these roles is substantive. For example, a Suzuki coupling is incomplete
without its organoboron partner and an esterification without its alcohol, whereas the
water of a hydrolysis may be left implicit.

\paragraph{Reactivity.}
Reactivity asks whether the listed reactants can actually give the target products.
There must be reasonable conditions, compatible with the listed reactants and products,
under which the listed reactants and any assumed auxiliary components give the target
products in isolable form. The target products need not be the major products. The
reaction must take place in a single step or a one-pot operation, without isolating an
intermediate.

\paragraph{Rationale.}
Both conditions follow from how proposals are produced and used in synthesis planning.
A step in a route must be executable from its listed starting materials, so a proposal
that omits a substantive partner cannot be kept. Retrosynthesis models, like the
reaction databases they are trained on, often omit routine
components~\citep{schneider2016uspto50k,wigh2024orderly}, which a chemist supplies
without comment. Requiring them would reject well-formed proposals, so auxiliary
components may be assumed. Proposals also leave the conditions to the chemist, so
reactivity depends on whether reasonable conditions exist, not on whether any are
stated. A route only needs to isolate the products of a step and carry them forward,
while the share of competing products depends on conditions that the proposal does not
fix, so the target products need not be the major ones. A sequence that isolates an
intermediate is two steps of a route, so a single step or a one-pot operation is
required. Finally, the assumptions are limited to what a chemist routinely supplies.
If any missing reactant, or any conditions however extreme, could be assumed, almost any
product would be reachable from almost any reactants, and an infeasible label would
carry no information.

\paragraph{What the labels assert.}
The labels record expert judgments under this criterion, which follows how a chemist
reviews a proposed step. They do not predict yield or certify experimental success.
A zero-yield record can therefore be labeled feasible if other reasonable conditions
could produce the target. \Cref{app:annotation} lists the permitted auxiliary roles and
gives the full decision procedure and annotation protocol.

%% file: sections/dataset.tex
\section{The \dataname Benchmark}
\label{sec:benchmark}

\dataname provides a common test of whether verifiers agree with experts
across different sources of reaction candidates. It contains \goldN{}
expert-labeled reactions, with \goldPos{} feasible and \goldNeg{} infeasible
under \cref{sec:task}.
The four sources connect practical screening, targeted diagnosis,
and evaluation of transfer (\cref{tab:benchmark_composition}).
Combining reactions encountered in practice with deliberate perturbations
allows us to test whether a verifier's strengths extend beyond one source.

\subsection{Four Reaction Sources}
\label{subsec:sources}

\input{tab/benchmark_composition}

\paragraph{Retrosynthesis proposals (RP).}
We collect proposals from eight single-step retrosynthesis models for targets
in ORDerly~\citep{wigh2024orderly}, a curated dataset derived from the Open
Reaction Database~\citep{kearnes2021ord}. The models span
template-based~\citep{chen2021localretro}, semi-template~\citep{chen2023g2retro},
and template-free approaches~\citep{zeng2024ualign,han2024editretro,
irwin2022chemformer,yu2024llasmol,baker2025larc}. Proposals are retained
whether or not they reproduce the recorded reactants. RP therefore tests
the proposals that verifiers must screen during planning, including valid
alternatives to the recorded reaction.

\paragraph{Zero-yield records (ZR).}
These are ORDerly reactions with a reported yield of zero. They were attempted
experimentally, but failure under the recorded conditions does not determine
feasibility under our criterion. On the contrary, chemists presumably attempted
them because they expected them to work. We therefore include them as potentially
feasible reactions and leave their labels unsettled until expert annotation.

\paragraph{LLM perturbations (LP).}
GPT-5.4~\citep{openai2026gpt54} and Claude Opus 4.6~\citep{anthropic2026claude46}
are instructed to edit recorded reactions to make them infeasible while keeping them
superficially coherent. This source challenges verifiers on identifying infeasible
edits among superficially coherent reactions.

\paragraph{Generated reactions (GR).}
We propose five methods for constructing negative candidates, building on
earlier uses of template application and reaction component replacement
\citep{segler20183nmcts,genheden2020filter,gainski2025retrogfn}.
The methods differ in what they change (\cref{fig:negative_methods}):
\begin{itemize}[leftmargin=1.5em, itemsep=1pt, topsep=2pt]
  \item \textbf{\negrafull (\negraabbr)} adds atoms at reactant reaction centers.
  \item \textbf{\negrrfull (\negrrabbr)} replaces the reactant set with a
    similar set from another recorded reaction.
  \item \textbf{\negncfull (\negncabbr)} chooses a different product
    disconnection and completes the resulting fragments into
    reactants~\citep{chen2023g2retro,baker2024rlsync}.
  \item \textbf{\negftfull (\negftabbr)} applies an alternative forward
    template to the recorded reactants.
  \item \textbf{\negetfull (\negetabbr)} applies a sequence of elementary
    reaction templates~\citep{joung2024elementary} to obtain a different product.
\end{itemize}
RA, RR and NC change the reactants while keeping the product. FT and ET change
the product while keeping the reactants. We sample the same number of
candidates from each method for annotation, retaining both feasible and infeasible examples.
The known generation methods allow failures to be traced to specific
constructions. Comparing GR with the other sources then tests whether gains
on these constructions extend to other reactions.
Applying the same methods to the curated ORDerly reactions also produces the
generated corpus used for training in \cref{sec:negatives}.
\Cref{app:curation,app:generation} give data collection and generation details.

\input{fig/negative_methods}

\subsection{Expert Annotation}
\label{subsec:annotation_quality}

Two chemists, a pharmacy faculty member and a chemistry postdoctoral researcher
with wet-laboratory experience, labeled reactions without seeing their sources
or model predictions.
The workflow combined guideline calibration, independent annotation and
joint review of disagreements, yielding \goldN{} expert-labeled benchmark
reactions.
The release includes recorded failure reasons and notes where available.
\Cref{app:annotation} gives the calibration procedure, inter-annotator
agreement and adjudication details.

The expert labels illustrate why source alone cannot determine feasibility:
79 of 93 zero-yield records are judged feasible, as are 33 of 289 generated
candidates intended to be negative. Expert labels therefore provide a
reference that neither recorded outcomes nor generation procedures supply
alone.

%% file: tab/benchmark_composition.tex
\begin{table}[t]
  \centering
  \caption{
    Composition of the \dataname benchmark.
    \emph{In practice}: reactions produced by retrosynthesis models or attempted
    in the laboratory. \emph{Controlled}: candidates obtained from a recorded
    reaction by a known negative candidate generation method.
    \Cref{tab:benchmark_breakdown} gives the per-model and per-method breakdown,
    including the fraction judged infeasible for each GR generation method.
  }
  \label{tab:benchmark_composition}
  \small
  \setlength{\tabcolsep}{3pt}
  \resizebox{\textwidth}{!}{%
  \begin{tabular}{@{}llccrrr@{}}
    \toprule
    \textbf{Source} & \textbf{Produced by} & \textbf{In practice}
      & \textbf{Controlled} & \textbf{Feas.} & \textbf{Infeas.}
      & \textbf{Total} \\
    \midrule
    Retrosynthesis proposals (RP) & Eight retrosynthesis models & \checkmark &  & \num{197} & \num{85} & \num{282} \\
    \addlinespace[2pt]
    Zero-yield records (ZR) & Experimental records & \checkmark &  & \num{79} & \num{14} & \num{93} \\
    \addlinespace[2pt]
    LLM perturbations (LP) & Two LLMs editing recorded reactions &  &  & \num{32} & \num{55} & \num{87} \\
    \addlinespace[2pt]
    Generated reactions (GR) & Five generation methods &  & \checkmark & \num{33} & \num{256} & \num{289} \\
    \midrule
    \textbf{Total} & & & & \num{341} & \num{410} & \num{751} \\
    \bottomrule
  \end{tabular}%
  }
\end{table}

%% file: fig/negative_methods.tex
\begin{figure}[t]
    \centering
    \includegraphics[width=\textwidth]{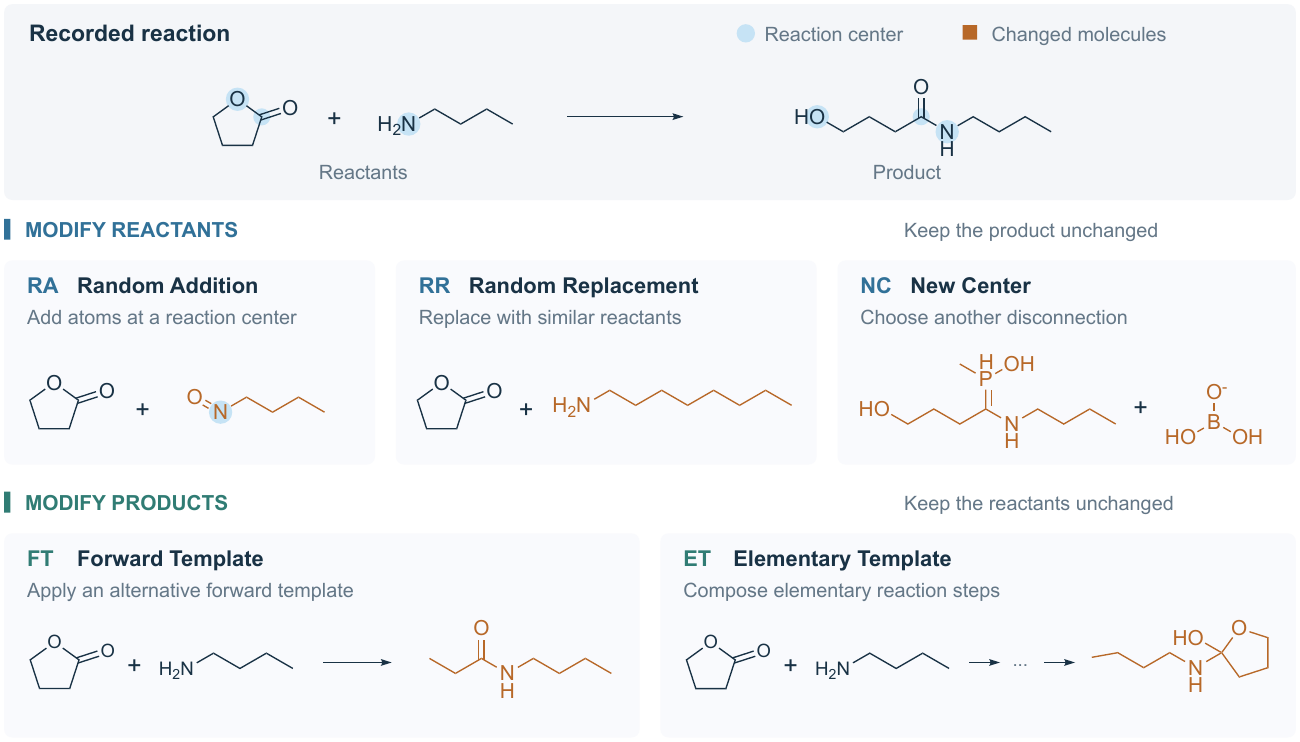}
    \caption{The five methods used to construct GR candidates, illustrated from a
    common parent reaction (top). RA, RR and NC change the reactants while keeping
    the product; FT and ET change the product while keeping the reactants.
    The methods generate intended negatives, whose feasibility is determined by
    expert annotation.}
    \label{fig:negative_methods}
\end{figure}

%% file: sections/protocol.tex
\section{Evaluation Protocol}
\label{sec:protocol}

\paragraph{Evaluated systems.}
We compare three groups of systems in the main results:
\begin{itemize}[leftmargin=1.5em, itemsep=2pt, topsep=2pt]
  \item \textbf{LLMs.} Claude Opus 5~\citep{anthropic2026claudeopus5} and
    GPT-6 Astra~\citep{openai2026gpt6astra} receive the reactants, products and
    feasibility criterion, and return a decision with an explanation.
  \item \textbf{Forward models.} Chemformer~\citep{irwin2022chemformer} and
    BARTSmiles~\citep{chilingaryan2024bartsmiles} are fine-tuned on recorded
    reactions. Top-10 inclusion accepts a proposal if its target matches one
    of the ten highest-ranked product predictions after canonicalization.
    Log-likelihood (LL) scores the supplied target by its mean per-token
    log-probability given the reactants.
  \item \textbf{Models trained on generated negatives.}
    \negname continues the BARTSmiles forward model with a loss that favors
    recorded reactions over generated negative candidates, and is scored by LL.
    \discname is trained to output a feasibility score directly.
    \gbdtname is a gradient-boosted tree using 58 chemical descriptors.
    All three use the training split of the generated corpus, with no benchmark
    reaction in their training data (\cref{app:splits}).
\end{itemize}
We also evaluate a rule-based baseline and the released filter from the
AiZynthFinder planning tool~\citep{genheden2020aizynthfinder},
with results in \cref{app:full_results}.

\paragraph{Metrics.}
The primary metric is AUROC. We also report balanced accuracy
(BAcc), the mean of the acceptance rate for feasible reactions and the rejection
rate for infeasible reactions. BAcc covers all systems, including the LLM and
top-10 decisions for which we extract no continuous score. Continuous-score
thresholds are fixed on held-out generated pairs before application to \dataname.
We report results for each source and their unweighted mean across the four sources.

\Cref{app:systems,app:protocol} provide implementation details and the
full evaluation protocol, including inference and training settings, model and
threshold selection, and statistical tests.

%% file: sections/results.tex
\section{Results}
\label{sec:results}

\input{tab/results_main}

\subsection{Performance Across Sources}
\label{subsec:overall}

We first ask whether verifiers agree with experts on each source, which determines how
reliably they can act as gatekeepers. \Cref{tab:results_main} reports AUROC and BAcc
within each source, together with their unweighted means.

\textbf{Frontier LLMs, given only the criterion, are strong candidate verifiers.}
Without examples from \dataname or parameter updates, Claude Opus 5 achieves
the highest four-source mean BAcc (68.0) and GPT-6 Astra reaches 65.7.
They also lead on the two edited sources: Claude on LP (76.1 BAcc) and GPT
on GR (62.7). Criterion prompting thus yields performance competitive with
dedicated verifiers, together with textual rationales for expert inspection.
However, they reject only 44.7--57.6\% of infeasible RP proposals, and each
rejects only 4 of the 14 infeasible ZR records (\cref{tab:app_rates}), leaving substantial
false acceptance on these sources. In addition, the binary outputs used here
do not support threshold tuning, while average API costs of US\$0.040--0.059
per reaction limit their practicality at scale (\cref{app:llm_baseline}).

\textbf{Forward models are the strongest verifiers of retrosynthesis proposals, but
reject most feasible edits of recorded reactions.}
BARTSmiles achieves the highest RP AUROC with likelihood scoring (78.9) and
BAcc with top-10 inclusion (73.1). Yet at the evaluated operating points,
the four forward-model readouts accept only 28.1--37.5\% of feasible LP
reactions and 12.1--33.3\% of feasible GR reactions, compared with
94.9--98.7\% on ZR (\cref{tab:app_rates}).
A possible explanation is that forward models are trained to reproduce recorded reactions,
so they may penalize LP and GR departures from recorded reaction patterns even when
the edits preserve feasibility.
Among the two readouts, we recommend LL in practice: it offers comparable
BAcc within each backbone, avoids the cost of beam search, and supports
threshold adjustment.

\textbf{Generated negatives give the highest mean AUROC when added to a forward model's
training, whereas classifiers trained on them directly have the lowest.}
In \cref{tab:results_main}, \negname achieves 75.2 mean AUROC, while
\discname and \gbdtname have lower means than the original forward models.
This contrast suggests that, in this comparison, generated negatives are
more useful for supplementing forward prediction than for training a
feasibility classifier directly. However,
the gains of \negname come mainly from LP and GR and do not extend to RP,
so its mean lead does not imply better screening of retrosynthesis proposals.
We further study the effects of negative supervision in \cref{sec:negatives}.

Overall, \textbf{no verifier leads across all sources}, indicating that agreement
with experts varies by candidate source. For screening retrosynthesis
proposals, BARTSmiles LL is a practical default among the evaluated systems,
while LLMs can provide rationales for selective expert review. Use \dataname
as an initial benchmark, then validate the model and calibrate any score
threshold on separate data representative of the intended
application.

\subsection{Performance by Generation Method}
\label{subsec:per_method}

\input{tab/results_per_method}

The differences across sources in \cref{subsec:overall} leave open whether
performance is consistent across the constructions within a source. We
examine this on GR, where each candidate's generation method is known.
\Cref{tab:results_per_method} compares each method's infeasible candidates
with the same 33 feasible GR reactions drawn from all five methods.

\textbf{Both forward models score below chance on NC negatives, ranking
infeasible candidates above feasible reference reactions.}
NC generates alternative product disconnections. Chemformer achieves 26.2
AUROC (95\% CI $[15,37]$) and BARTSmiles achieves 32.5 ($[20,45]$).
Both intervals lie entirely below 50, indicating an inversion of the
desired ranking on this comparison. The same models achieve 83.5--94.9
AUROC on FT and ET, which change the recorded product, and 79.8--83.5 on
RR, which replaces the reactant set. Both are near chance on RA, which
makes local reactant edits. Evaluation on FT, ET or RR alone would miss
this failure on alternative disconnections.

\textbf{\negname improves NC and RA scores but reduces RR performance
relative to BARTSmiles LL.}
NC increases from 32.5 to 94.2 and RA from 46.1 to 72.7, while RR decreases
from 79.8 to 53.4. The higher overall GR AUROC thus combines substantial
gains and losses across methods. The discriminative verifier exhibits a
similar profile and scores below \negname on four of the five methods.

To include systems that return only decisions, we also examine rejection
rates (\cref{tab:app_per_method_rates}). Claude rejects at least 83\% of
infeasible candidates from each method, but also rejects two-thirds of the
feasible GR reactions. Its high negative rejection rates must therefore be
considered together with its rejection of feasible alternatives.

These comparisons show that source-level scores can conceal systematic
failures on particular constructions. A separate analysis in
\cref{app:ranking} shows that reaction-level performance also need not
preserve expert rankings of retrosynthesis models.

For reaction-level verification, the selective gains of \negname leave a
training question open: can generated negatives improve agreement with
experts beyond specific constructions, especially on RP? We address this
with matched training comparisons in \cref{sec:negatives}.

%% file: tab/results_main.tex
\begin{table}[t]
  \centering
  \caption{
    Results on \dataname by source (feasible/infeasible counts in parentheses).
    Avg.\ denotes the four-source mean. Dashes indicate no continuous score.
    Best values per column are bold.
    Counts describe the benchmark; one unparseable Claude response on LP is excluded
    from that model's metrics.
  }
  \label{tab:results_main}
  \small
  \setlength{\tabcolsep}{2.3pt}
  \begin{tabular}{@{}l cc cc cc cc cc@{}}
    \toprule
    & \multicolumn{2}{c}{RP {\scriptsize(197/85)}} & \multicolumn{2}{c}{ZR {\scriptsize(79/14)}} & \multicolumn{2}{c}{LP {\scriptsize(32/55)}} & \multicolumn{2}{c}{GR {\scriptsize(33/256)}} & \multicolumn{2}{c}{Avg.} \\
    \cmidrule(lr){2-3}\cmidrule(lr){4-5}\cmidrule(lr){6-7}\cmidrule(lr){8-9}\cmidrule(l){10-11}
    System & AUROC & BAcc & AUROC & BAcc & AUROC & BAcc & AUROC & BAcc & AUROC & BAcc \\
    \midrule
    \multicolumn{11}{@{}l}{\textit{Large language models, criterion in the prompt}} \\
    Claude Opus 5 & -- & 69.9 & -- & 64.3 & -- & \textbf{76.1} & -- & 61.8 & -- & \textbf{68.0} \\
    GPT-6 Astra & -- & 66.8 & -- & 63.7 & -- & 69.6 & -- & \textbf{62.7} & -- & 65.7 \\
    \midrule
    \multicolumn{11}{@{}l}{\textit{Forward-model proxies}} \\
    Chemformer (top-10) & -- & 70.6 & -- & 67.2 & -- & 63.3 & -- & 50.1 & -- & 62.8 \\
    Chemformer (LL) & 76.7 & 71.6 & \textbf{82.4} & 70.2 & 64.5 & 63.8 & 67.0 & 49.8 & 72.6 & 63.9 \\
    BARTSmiles (top-10) & -- & \textbf{73.1} & -- & \textbf{74.4} & -- & 59.0 & -- & 57.1 & -- & 65.9 \\
    BARTSmiles (LL) & \textbf{78.9} & 71.5 & 79.5 & 72.5 & 66.1 & 60.4 & 66.9 & 53.4 & 72.8 & 64.4 \\
    \midrule
    \multicolumn{11}{@{}l}{\textit{Trained on our generated negative candidates}} \\
    \negname & 77.0 & 67.1 & 80.8 & 63.7 & 69.4 & 66.1 & \textbf{73.5} & 54.9 & \textbf{75.2} & 62.9 \\
    \discname & 67.8 & 62.5 & 79.0 & 60.1 & 66.0 & 54.8 & 66.0 & 56.6 & 69.7 & 58.5 \\
    \gbdtname & 70.3 & 64.0 & 72.2 & 56.3 & \textbf{75.4} & 70.2 & 64.7 & 62.4 & 70.7 & 63.2 \\
    \bottomrule
  \end{tabular}
\end{table}

%% file: tab/results_per_method.tex
\begin{table}[t]
  \centering
  \caption{
    AUROC on GR by the method that produced each candidate, with DeLong 95\% confidence
    intervals. Each column ranks the infeasible candidates of one method (count in
    parentheses) against the same \num{33} feasible GR candidates, so 50 is chance and a
    value below 50 ranks infeasible candidates above feasible ones. The measure is
    threshold-free, so it covers only systems that return a score; rejection rates at
    each operating point, including those of the language models and the top-10
    readouts, are in \cref{tab:app_per_method_rates}.
  }
  \label{tab:results_per_method}
  \small
  \setlength{\tabcolsep}{4pt}
  \begin{tabular}{@{}l ccc cc@{}}
    \toprule
    & \multicolumn{3}{c}{Reactants changed} & \multicolumn{2}{c}{Products changed} \\
    \cmidrule(lr){2-4}\cmidrule(l){5-6}
    System & \negraabbr{\scriptsize(50)} & \negrrabbr{\scriptsize(53)} & \negncabbr{\scriptsize(54)} & \negftabbr{\scriptsize(44)} & \negetabbr{\scriptsize(55)} \\
    \midrule
    \multicolumn{6}{@{}l}{\textit{Forward-model proxies}} \\
    Chemformer (LL) & 44.7 {\scriptsize\color{gray}[32, 57]} & 83.5 {\scriptsize\color{gray}[75, 93]} & 26.2 {\scriptsize\color{gray}[15, 37]} & 87.5 {\scriptsize\color{gray}[79, 96]} & 94.9 {\scriptsize\color{gray}[90, 100]} \\
    BARTSmiles (LL) & 46.1 {\scriptsize\color{gray}[33, 60]} & 79.8 {\scriptsize\color{gray}[70, 90]} & 32.5 {\scriptsize\color{gray}[20, 45]} & 83.5 {\scriptsize\color{gray}[74, 93]} & 93.7 {\scriptsize\color{gray}[88, 99]} \\
    \midrule
    \multicolumn{6}{@{}l}{\textit{Trained on our generated negative candidates}} \\
    \negname & 72.7 {\scriptsize\color{gray}[62, 84]} & 53.4 {\scriptsize\color{gray}[40, 66]} & 94.2 {\scriptsize\color{gray}[90, 99]} & 67.7 {\scriptsize\color{gray}[55, 81]} & 78.1 {\scriptsize\color{gray}[67, 89]} \\
    \discname & 63.9 {\scriptsize\color{gray}[51, 77]} & 46.5 {\scriptsize\color{gray}[33, 60]} & 78.0 {\scriptsize\color{gray}[66, 90]} & 68.5 {\scriptsize\color{gray}[56, 81]} & 72.9 {\scriptsize\color{gray}[61, 85]} \\
    \gbdtname & 50.4 {\scriptsize\color{gray}[36, 65]} & 45.6 {\scriptsize\color{gray}[32, 59]} & 87.0 {\scriptsize\color{gray}[79, 95]} & 67.8 {\scriptsize\color{gray}[56, 80]} & 72.0 {\scriptsize\color{gray}[60, 84]} \\
    \bottomrule
  \end{tabular}
\end{table}

%% file: sections/negatives.tex
\section{Training with Generated Negatives}
\label{sec:negatives}

The preceding analyses locate failures tied to candidate construction.
We now ask whether generated negatives can improve verification across
sources, particularly on RP. Using generated training pairs that share no reaction
with the benchmark (\cref{app:splits}), we first isolate the effect of
adding negative supervision (\cref{subsec:neg_controlled}), then vary its
generation method (\cref{subsec:neg_permethod}). Each setting uses a single
training run, so the observed gains remain exploratory.

\subsection{Adding Negative Supervision}
\label{subsec:neg_controlled}

\input{tab/neg_controlled}

The comparison in \cref{subsec:overall} does not isolate negative
supervision from continued forward training. To separate their effects,
we compare \negname with a control that starts from the same BARTSmiles
forward model and uses the same positive reactions and training schedule.
Both runs continue for one epoch over \num{800000} pairs
(\cref{app:controlled_pair}).
The control optimizes only the forward-prediction loss.
\negname adds a preference loss that ranks each recorded reaction above its
generated candidate, using the same mean per-token log-likelihood as at
evaluation. This preserves forward-prediction training while introducing
supervision from the negative candidates.

\textbf{The observed mean gain is concentrated outside retrosynthesis
proposals.}
The four-source mean AUROC increases from 71.4 for the continued-training
control to 75.2 with negatives (\cref{tab:neg_controlled}). The starting model
scores 72.8, so the gain is not reproduced by the additional epoch of
positive-only training.
Scores increase on ZR, LP and GR, but decrease slightly on RP.

Repeating the comparison with \num{2400000} pairs gives the same directions
of change (\cref{tab:app_full_auroc}). None of the within-source differences
survives Holm correction (\cref{app:stats}). Because these runs use a mixture
of generated negatives, they leave open whether the lack of RP improvement
is shared across generation methods or depends on the mixture.

\subsection{Choosing the Generation Method}
\label{subsec:neg_permethod}

\input{tab/neg_permethod}

To examine this possibility, we vary the generation method supplying the
training pairs. This complements \cref{subsec:per_method}, which varied
test constructions for fixed verifiers: here we measure how training on
different negatives transfers across sources. We compare each method
alone, an equal mixture of all five, and a control without negatives.
All seven runs share the initialization, positive stream for forward
prediction and training schedule. Each run with negatives uses
\num{800000} pairs; only the pairs supplied to the preference loss change.
We evaluate all runs at their final training step under the pre-registered
design (\cref{app:prereg}).

\textbf{The observed rankings of generation methods differ between GR and RP.}
In these runs, NC raises GR AUROC by 12.1 points but lowers RP by 2.8 points
(\cref{tab:neg_permethod}). RR raises mean AUROC by 3.2 points, followed by
the mixture at 2.3 points, while ET leaves the mean nearly unchanged.
Both RR and RA slightly improve RP over the control.
The equal mixture has a lower mean than RR and also lowers RP by 2.3 points.

Only the FT decline of 6.4 points in mean AUROC is significant after
Holm correction ($p_{\mathrm{adj}} = 0.001$). The RR gain has
$p_{\mathrm{adj}} = 0.08$, leaving its benefit over the control inconclusive.
FT also has the lowest estimated negative purity---the fraction of candidates
experts judge infeasible---at 77.2\%, compared with about 93\% for RR, NC and ET
(\cref{tab:benchmark_breakdown}).
FT's decline therefore cannot be attributed to its construction independently
of label noise. Together with the differing outcomes of RR, NC and ET,
this motivates examining label reliability alongside other properties of
the generated candidates.

\subsection{Hypotheses for Improving Generalization}
\label{subsec:neg_discussion}

The method comparison shows that gains on generated constructions need not
carry over to RP, but leaves the reasons for this limited transfer open.
We therefore combine the training results with expert label audits and
the descriptor-classifier analysis to formulate three hypotheses for
improving negative-data design. Each requires a separate test.

\textbf{Validating negative labels may improve transfer by preserving
feasible alternatives.}
FT combines the lowest estimated negative purity with the largest
performance decline. Since generated candidates can remain feasible,
assigning them negative labels can penalize valid alternatives to the
recorded reaction. This motivates expert audits of recurring labeling
errors, followed by comparisons that filter or down-weight uncertain
candidates while holding the generation method fixed.

\textbf{Reducing simple descriptor differences may encourage more
transferable learning.}
Label reliability alone may not account for the outcomes: RR, NC and ET
have comparable estimated purities but different observed transfer patterns.
Among these three, RR candidates are the hardest for the descriptor
classifier to separate from recorded reactions (\cref{tab:app_shallow_sep}),
and RR gives the largest observed mean gain. This association motivates
testing generation methods that match recorded reactions and negative
candidates on simple molecular descriptors while controlling label purity.

\textbf{Targeting errors in model proposals may improve transfer more than
simply mixing generation methods.}
Beyond label reliability and descriptor differences, useful negatives
should address the errors encountered in model proposals. The NC and mixture
runs both improve GR while lowering RP, motivating a test of candidates
targeted to those errors. Expert-reviewed proposals from a separate
development set could identify recurring errors and guide perturbations
and method weights. Held-out proposals would then test whether these
targeted choices improve transfer.

%% file: tab/neg_controlled.tex
\begin{table}[t]
  \centering
  \caption{
    One further epoch of training without and with generated negative candidates,
    with matched positive reactions and training schedules. AUROC by source,
    log-likelihood readout. \emph{w/ neg.} is \negname of
    \cref{tab:results_main,tab:results_per_method}.
  }
  \label{tab:neg_controlled}
  \small
  \setlength{\tabcolsep}{10pt}
  \begin{tabular}{@{}l ccccc@{\hspace{16pt}}}
    \toprule
    & RP & ZR & LP & GR & Avg. \\
    \midrule
    BARTSmiles & \makebox[2.3em][r]{78.9} & \makebox[2.3em][r]{79.5} & \makebox[2.3em][r]{66.1} & \makebox[2.3em][r]{66.9} & \makebox[2.3em][r]{72.8} \\
    $+$ w/o neg. & \makebox[2.3em][r]{77.8}\rlap{\,\textcolor{gray}{\tiny$-1.1$}} & \makebox[2.3em][r]{76.5}\rlap{\,\textcolor{gray}{\tiny$-3.0$}} & \makebox[2.3em][r]{66.8}\rlap{\,\textcolor{gray}{\tiny$+0.7$}} & \makebox[2.3em][r]{64.5}\rlap{\,\textcolor{gray}{\tiny$-2.4$}} & \makebox[2.3em][r]{71.4}\rlap{\,\textcolor{gray}{\tiny$-1.4$}} \\
    $+$ w/ neg. & \makebox[2.3em][r]{77.0}\rlap{\,\textcolor{gray}{\tiny$-1.8$}} & \makebox[2.3em][r]{80.8}\rlap{\,\textcolor{gray}{\tiny$+1.4$}} & \makebox[2.3em][r]{69.4}\rlap{\,\textcolor{gray}{\tiny$+3.4$}} & \makebox[2.3em][r]{73.5}\rlap{\,\textcolor{gray}{\tiny$+6.7$}} & \makebox[2.3em][r]{75.2}\rlap{\,\textcolor{gray}{\tiny$+2.4$}} \\
    \midrule
    $\Delta$ (w/ $-$ w/o) & \makebox[2.3em][r]{$-0.7$} & \makebox[2.3em][r]{$+4.3$} & \makebox[2.3em][r]{$+2.6$} & \makebox[2.3em][r]{$+9.0$} & \makebox[2.3em][r]{$+3.8$} \\
    \bottomrule
  \end{tabular}
\end{table}

%% file: tab/neg_permethod.tex
\begin{table}[t]
  \centering
  \caption{
    The same recipe with the negatives restricted to the candidates of one
    generation method, and with an equal share of all five. AUROC by source,
    log-likelihood readout. Grey: difference from \emph{w/o neg.}, which is a
    separately trained control using the same positive-only recipe as
    \cref{tab:neg_controlled} and the shared positive stream of this experiment
    (\cref{app:prereg}).
  }
  \label{tab:neg_permethod}
  \small
  \setlength{\tabcolsep}{10pt}
  \begin{tabular}{@{}l ccccc@{\hspace{16pt}}}
    \toprule
    & RP & ZR & LP & GR & Avg. \\
    \midrule
    w/o neg. & \makebox[2.3em][r]{77.6} & \makebox[2.3em][r]{\textbf{79.0}} & \makebox[2.3em][r]{65.5} & \makebox[2.3em][r]{64.8} & \makebox[2.3em][r]{71.7} \\
    \addlinespace[1pt]
    w/ only \negraabbr neg. & \makebox[2.3em][r]{78.6}\rlap{\,\textcolor{gray}{\tiny$+1.0$}} & \makebox[2.3em][r]{76.8}\rlap{\,\textcolor{gray}{\tiny$-2.3$}} & \makebox[2.3em][r]{67.3}\rlap{\,\textcolor{gray}{\tiny$+1.8$}} & \makebox[2.3em][r]{68.0}\rlap{\,\textcolor{gray}{\tiny$+3.2$}} & \makebox[2.3em][r]{72.7}\rlap{\,\textcolor{gray}{\tiny$+0.9$}} \\
    w/ only \negrrabbr neg. & \makebox[2.3em][r]{\textbf{78.9}}\rlap{\,\textcolor{gray}{\tiny$+1.3$}} & \makebox[2.3em][r]{\textbf{79.0}}\rlap{\,\textcolor{gray}{\tiny$+0.0$}} & \makebox[2.3em][r]{\textbf{74.4}}\rlap{\,\textcolor{gray}{\tiny$+8.9$}} & \makebox[2.3em][r]{67.4}\rlap{\,\textcolor{gray}{\tiny$+2.6$}} & \makebox[2.3em][r]{\textbf{74.9}}\rlap{\,\textcolor{gray}{\tiny$+3.2$}} \\
    w/ only \negncabbr neg. & \makebox[2.3em][r]{74.8}\rlap{\,\textcolor{gray}{\tiny$-2.8$}} & \makebox[2.3em][r]{76.4}\rlap{\,\textcolor{gray}{\tiny$-2.6$}} & \makebox[2.3em][r]{67.4}\rlap{\,\textcolor{gray}{\tiny$+1.9$}} & \makebox[2.3em][r]{\textbf{76.9}}\rlap{\,\textcolor{gray}{\tiny$+12.1$}} & \makebox[2.3em][r]{73.9}\rlap{\,\textcolor{gray}{\tiny$+2.2$}} \\
    w/ only \negftabbr neg. & \makebox[2.3em][r]{70.7}\rlap{\,\textcolor{gray}{\tiny$-6.9$}} & \makebox[2.3em][r]{76.2}\rlap{\,\textcolor{gray}{\tiny$-2.8$}} & \makebox[2.3em][r]{62.2}\rlap{\,\textcolor{gray}{\tiny$-3.4$}} & \makebox[2.3em][r]{52.4}\rlap{\,\textcolor{gray}{\tiny$-12.4$}} & \makebox[2.3em][r]{65.4}\rlap{\,\textcolor{gray}{\tiny$-6.4$}} \\
    w/ only \negetabbr neg. & \makebox[2.3em][r]{77.3}\rlap{\,\textcolor{gray}{\tiny$-0.3$}} & \makebox[2.3em][r]{77.3}\rlap{\,\textcolor{gray}{\tiny$-1.7$}} & \makebox[2.3em][r]{68.5}\rlap{\,\textcolor{gray}{\tiny$+3.0$}} & \makebox[2.3em][r]{64.5}\rlap{\,\textcolor{gray}{\tiny$-0.3$}} & \makebox[2.3em][r]{71.9}\rlap{\,\textcolor{gray}{\tiny$+0.2$}} \\
    \addlinespace[1pt]
    w/ all-method neg. & \makebox[2.3em][r]{75.3}\rlap{\,\textcolor{gray}{\tiny$-2.3$}} & \makebox[2.3em][r]{77.3}\rlap{\,\textcolor{gray}{\tiny$-1.7$}} & \makebox[2.3em][r]{71.0}\rlap{\,\textcolor{gray}{\tiny$+5.5$}} & \makebox[2.3em][r]{72.5}\rlap{\,\textcolor{gray}{\tiny$+7.7$}} & \makebox[2.3em][r]{74.0}\rlap{\,\textcolor{gray}{\tiny$+2.3$}} \\
    \bottomrule
  \end{tabular}
\end{table}

%% file: sections/conclusion.tex
\section{Conclusion}
\label{sec:conclusion}

\dataname provides an expert reference for reaction feasibility verification across
multiple candidate sources. Its results reveal complementary strengths: criterion-prompted
LLMs are competitive with dedicated verifiers, while forward models lead on retrosynthesis
proposals yet fail systematically on some constructed candidates. No verifier leads
across all sources, and strong aggregate performance can conceal these failures.
Verifiers should therefore be selected and calibrated on data representative of their
intended screening task.

Our matched training experiments show how the benchmark can guide improvement:
adding a mixture of generated negatives raises mean AUROC without improving performance
on model proposals, while transfer varies with the generation method.
This makes performance on those proposals a central test for negative-data design.
The results motivate testing whether auditing negative labels, controlling descriptor
differences and targeting recurring proposal errors improve transfer.
These directions remain hypotheses given the benchmark's limited coverage, judgment
uncertainty and single-run training comparisons (\cref{app:limitations}). Together,
the benchmark and accompanying corpus support developing verifiers whose improvements
are measured against expert judgments on the reactions they must screen.

%% file: sections/acknowledgments.tex
\section*{Acknowledgments}

This research was supported in part by NLM R01LM014385 and NSF \#2435819. The content is
solely the responsibility of the authors and does not necessarily represent the official
views of the National Institutes of Health or the National Science Foundation.

%% file: appendices/related_work.tex
\section{Related Work, Extended}
\label{app:related}

This appendix expands \cref{sec:related} and compares the resources it mentions
(\cref{tab:app_related}).

\paragraph{Filters trained on constructed negatives.}
\citet{segler20183nmcts} introduced an in-scope filter: a neural classifier trained to
distinguish recorded reactions from constructed ones. Its \num{100} million training
negatives were generated by forward rule application and negative sampling;
\citet{genheden2020filter}, who reimplemented the filter, describe the two
strategies as taking the further products that the recorded template yields on the
recorded reactants (\num{30} million) and applying a randomly chosen template to a random
reactant set (\num{70} million). AiZynthFinder ships a filter built on the same
principle, and~\citet{genheden2020filter} compare three ways of drawing the negatives,
all template-based and all keeping the recorded reactants: applying the recorded template
to give a further product, applying a random template, and applying a template suggested
by a recommender. RetroGFN's reaction feasibility model is trained on negatives from two
procedures, applying forward templates to the recorded reactants to obtain a product that
was not recorded, and swapping the recorded product for another product of high Tanimoto
similarity~\citep{gainski2025retrogfn}. The same paper separately builds a set of hard
negatives in the way our \negrrabbr method does, pairing a template-generated product
with reactant sets of high Tanimoto similarity to the ones it was generated from, and
uses that set to estimate how often a forward model accepts an infeasible reaction, which
is the measurement \cref{sec:results} makes against expert labels instead. RetroTrim assembles an ensemble of
scorers, some learned and some derived from reaction databases, and shows that different
scorers catch different classes of hallucinated reaction. It evaluates reaction scores
against expert labels and complete routes for 32 targets~\citep{sadowski2025retrotrim}.
Its reaction-level evaluation treats Safe Bet as positive and Rather Not and Nonsense
as negative, excluding Worthwhile reactions. Filters in this line of work have been
assessed on constructed test reactions and through their effects on downstream search;
\citet{genheden2020filter} also test transfer between negative generation methods.

\paragraph{Evaluation based on precedent.}
\citet{zagribelnyy2026chemcensor} argue that scoring single-step retrosynthesis by exact
match against one recorded answer penalizes valid alternatives, and propose instead to
extract a proposed reaction's reaction center and its surrounding functional-group
context and check both against a relational database of reported reactions, assigning a
score of zero when no precedent is found. They release CREED, a corpus of \num{22712526}
reactions for \num{1493715} products generated from roughly 3,000 expert-written
bidirectional templates. Precedent-based scoring is complementary to our
expert judgments, which can accept a sound proposal without a matching
recorded precedent. CREED supplies plausible generated reactions for
training. \dataname combines feasible and infeasible expert-labeled
candidates across four sources to evaluate verifiers and the transfer
of generated-negative training.

\paragraph{Negative reactions from experiment.}
\citet{jiang2021smiles} trained on failed reactions recorded in a private industrial
laboratory notebook, a resource that is not public.
\citet{zhong2025feasibilityHTE} released high-throughput experimentation data for
acid--amine couplings with Bayesian deep learning, including observed failures under
specified conditions for a single reaction class. DORA-XGB builds an enzymatic
reaction feasibility classifier, taking unreported reactions as
infeasible~\citep{chainani2025dora}; the assumption is strong and the domain is
retrobiosynthesis. Zero-yield records are plentiful in public
databases~\citep{kearnes2021ord,wigh2024orderly} but report an outcome under particular
conditions, so they cannot by themselves establish that a transformation is impossible.
This is why our ZR source enters the benchmark as candidates for annotation rather than
as negatives.

\paragraph{Feasibility from first principles.}
\citet{rappoport2019predicting} explore reaction pathways with heuristically aided
quantum chemistry, and~\citet{sigmund2025qmmlfeasibility} combine quantum-chemical
descriptors with machine learning to predict feasibility and selectivity for one reaction
type. Mechanistic datasets support models that reproduce elementary
steps~\citep{joung2024elementary}. Mechanistic and energetic evidence from these
approaches could complement expert review of difficult proposals in future extensions.

\paragraph{Adjacent tasks.}
Yield prediction asks how much product forms under stated conditions rather than whether
a proposal is well formed, and its own difficulties are
documented~\citep{voinarovska2023yield}. Forward reaction prediction asks what products
form~\citep{schwaller2019molecular,irwin2022chemformer,chilingaryan2024bartsmiles}, and is
repurposed for verification by the round-trip
convention~\citep{schwaller2019evaluation,maziarz2025syntheseus} that
\cref{sec:results} evaluates.

\begin{table}[h]
  \centering
  \caption{
    Selected resources related to reaction feasibility. \emph{Expert labels}: whether
    individual reactions carry feasibility judgments by chemists. \emph{Negatives}:
    whether the resource includes experimental failures or reactions labeled infeasible.
  }
  \label{tab:app_related}
  \small
  \setlength{\tabcolsep}{4pt}
  \begin{tabular}{@{}>{\raggedright\arraybackslash}p{0.26\textwidth} >{\raggedright\arraybackslash}p{0.17\textwidth} c c >{\raggedright\arraybackslash}p{0.21\textwidth}@{}}
    \toprule
    Resource & Scope & Expert labels & Negatives & Source of negatives \\
    \midrule
    \citet{jiang2021smiles}      & One laboratory & --- & \checkmark & failed experiments (private) \\
    \citet{zhong2025feasibilityHTE} & Acid--amine couplings & --- & \checkmark & high-throughput experiments \\
    DORA-XGB~\citep{chainani2025dora} & Enzymatic & --- & \checkmark & unreported reactions \\
    RetroGFN~\citep{gainski2025retrogfn} & Small molecule & --- & \checkmark & templates, product swap \\
    RetroTrim~\citep{sadowski2025retrotrim} & Drug-like targets & \checkmark & \checkmark & expert-rated proposals \\
    CREED~\citep{zagribelnyy2026chemcensor} & Small molecule & --- & --- & --- \\
    \dataname (ours) & Small molecule & \checkmark & \checkmark & four sources (\cref{sec:benchmark}) \\
    \bottomrule
  \end{tabular}
\end{table}

%% file: appendices/annotation.tex
\section{Feasibility Criterion and Annotation Protocol}
\label{app:annotation}

This appendix states the criterion of \cref{sec:task} in full and gives its rationale
(\cref{app:criterion,app:rationale}), specifies what annotators record
(\cref{app:procedure,app:reasons,app:feasible_notes}), and describes how the evaluation
set was annotated (\cref{app:annotators,app:calibration,app:agreement,app:screens}).

\subsection{Criterion}
\label{app:criterion}

For each reaction, the annotator decides whether, under some reasonable conditions, the
listed reactants can give the listed products in isolable form, in a single step or a
credible one-pot operation.
Only the reactant and product lists are given; there are no separate fields for
reagents, conditions or yield. Any species explicitly listed on the reactant side
remains part of the proposal.
The judgment combines the two conditions of \cref{sec:task}, which are applied as
follows.

\paragraph{Completeness.}
Every species on the left-hand side counts as a reactant, whether or not a reaction
database would record it as a reagent; what matters is whether an omitted species is
essential. An omitted species may be assumed only if it acts in one of the permitted roles
of \cref{tab:auxiliary_boundary}, within the scope stated there. Such a species is
\emph{auxiliary}. Anything else is \emph{substantive} and may not be assumed. A
transformation evidently intended as a Suzuki coupling is not a complete proposal without
its organoboron partner, and an esterification written without its alcohol is not one
either. Status follows from the role a species plays
here, not from its name, size or customary classification
(\cref{tab:boundary_examples}), so methanol is auxiliary as a solvent and substantive as
the source of a methyl ester. A missing substantive reactant makes a reaction infeasible
even when the intended transformation is recognizable.

A proposal that is complete but for a permitted auxiliary carries a
\texttt{missing\_auxiliary\_reagent} note naming the species and its role
(\cref{app:feasible_notes}).

\input{tab/auxiliary_boundary}
\input{tab/boundary_examples}

\paragraph{The test is structural, not stoichiometric.}
The test cannot be implemented by comparing element counts, which fail in both
directions.
Counts can balance although a substantive partner is missing, because leaving groups
and other listed species routinely supply the same elements.
In one proposal from our pool, a substrate and the coupling reagent HATU are offered as
giving a product that bears a 2-picolylamine group; the element counts balance, since
HATU supplies carbon, nitrogen and oxygen, yet no listed species contains a
picolylamine fragment, so the amine partner is missing.
Conversely, counts can fail to balance although nothing is missing: the test concerns
only atoms retained in the products, so omitted by-products, unspecified stoichiometry
and differences in hydrogen count are not defects, provided a credible hydrogen source
exists among the permitted additions.
Annotators are therefore asked where each fragment of the products comes from,
structurally, rather than whether the formula balances.

\paragraph{Reactivity.}
The annotator may assume any reasonable conditions that are not stated, including
solvent, temperature, time, atmosphere and order of addition, but may not replace,
modify or disregard a listed species.
A reaction satisfies reactivity if, under some such conditions, the listed reactants,
completed only by auxiliary species, give the listed products in a form that can be
isolated.
They may form alongside other products, which need not be listed. What is required
is that the listed products can be obtained and separated.
Assumed conditions must remain compatible with the reactants and products;
hypothetical extreme conditions cannot rescue an otherwise unsupported transformation.
The reaction must take place in a single step or a credible one-pot operation, which may
include sequential addition and the \emph{in situ} formation of a reactive species;
workup does not count as a separate step, whereas the isolation of an intermediate does.
A reaction fails this condition only if no reasonable conditions would yield the
listed products.

\subsection{Rationale and scope}
\label{app:rationale}

\paragraph{Why an explicit definition is needed.}
Feasibility is used in the synthesis-planning literature in at least four senses: a
reaction may be called feasible because it is recorded in a database, recoverable by a
forward model from its reactants, precedented in its reaction center and
functional-group context, or executable in a laboratory.
These senses agree on routine transformations but diverge on the cases that matter
most for evaluation: an unprecedented but sound disconnection may be executable,
whereas a precedented transformation applied to an incompatible substrate is not.
A benchmark must therefore state which sense it measures.
Ours is the validity of a retrosynthesis proposal, judged by experts under the rules of
this appendix.

\paragraph{Why the two conditions are defined this way.}
The criterion reflects chemical practice and the requirements of synthesis planning.
\emph{Chemical practice.}
A chemist reviewing a proposed step reads everything on the left-hand side as
reactants and judges an omission by whether the omitted species is essential: a
solvent, water or a routine additive is supplied without comment, a partner that builds
part of the product is not.
Because a proposal specifies no conditions, the chemist likewise asks whether
reasonable conditions exist, not whether particular conditions are stated.
\emph{Application.}
In retrosynthesis planning, a proposal that omits a routine component still describes a
usable step and should be expanded, whereas one that omits the coupling partner does
not describe a step at all and should be pruned.
A step is useful in a route if its product can be isolated and carried forward, and how
the material divides among competing products depends on conditions that a proposal does
not fix, so reactivity asks only that the listed products can be isolated.
\emph{Informativeness.}
If any unlisted species could be assumed, or any conditions however extreme, almost any
product would be reachable from almost any reactants, and a negative label would carry
no information.

\paragraph{What the label asserts.}
A positive label requires more than a recognizable transformation, as the completeness
condition makes explicit.
A negative label denotes a defect in the proposal as stated; it does not assert that
the transformation is impossible under every conceivable condition, a claim that expert
annotation cannot establish.
Proposal feasibility is also distinct from \emph{experimental} feasibility, which
additionally requires that suitable conditions be identified and that the reaction
proceed in acceptable yield in practice.
Retrosynthesis proposals carry no conditions, and holding a model that reads them
responsible for experimental outcomes would penalize it for information it is not
given.
The same reasoning applies to experimental records: a reported success provides
evidence that suitable conditions exist, whereas a reported failure records an
unsuccessful attempt without settling whether other reasonable conditions could work.
Records of failed experiments therefore enter the evaluation set as candidates and are
labeled under the same criterion as every other source.

\paragraph{Independence from the source of a reaction.}
The criterion refers only to the listed reactants and products, not to how the
reaction was produced.
It therefore applies unchanged to proposals from retrosynthesis models, reactions
edited by language models, experimental records and generated candidates, which is what
allows performance to be compared across the sources of the evaluation set.

\subsection{Decision procedure and recorded fields}
\label{app:procedure}

For each proposal, the annotator records a label (\emph{feasible}, \emph{infeasible} or
\emph{uncertain}); one or more failure reasons if the label is infeasible
(\cref{app:reasons}); one or more notes if it is feasible (\cref{app:feasible_notes});
a difficulty rating; and a free-text note, which is required for \emph{uncertain} and
recommended otherwise.
\emph{Uncertain} is not a softened negative.
It is reserved for proposals on which the annotator genuinely cannot decide, with the
reason stated; an annotator who leans towards one label records that label.

Every proposal is taken through the ordered checks of \cref{tab:decision_procedure}.
Checks~1--3 take precedence: if one fails, its reason is recorded and later checks
are not applied. Otherwise, checks~4--6 are assessed and all applicable reasons may
be recorded. A proposal that passes all six is feasible.
This precedence is part of the protocol.
A defective proposal can often be described in more than one way---a missing amine
partner, for instance, as an absent reactant or as a transformation without mechanistic
basis---and without a fixed precedence, two annotators who agree that a proposal is
infeasible may still record different reasons.
Fixing the order makes the recorded reason reproducible.
Two further rules apply.
The reasons at check~3 are alternatives: \texttt{missing\_substantive\_reactant}
applies when the intended transformation is recognizable and its only defect is an
unlisted substantive partner, and \texttt{reactants\_cannot\_give\_product} applies
otherwise; if both are selected, only the latter is kept.
Reasons from checks~4--6 may be recorded together when more than one applies.

\input{tab/decision_procedure}

\subsection{Failure reasons}
\label{app:reasons}

\Cref{tab:failure_reasons} defines the seven reasons.
They map onto the two conditions of \cref{sec:task}:
\texttt{missing\_substantive\_reactant} records a failure of completeness, and each of
the other six records a failure of reactivity.
Two distinctions guide the choice of reason.
First, \texttt{reactivity\_conflict} applies when the intended transformation is sound
but a competing process prevents the products from being isolated, whereas \texttt{reactants\_cannot\_give\_product}
applies when the transformation has no valid basis to begin with.
Second, competing outcomes do not by themselves make a proposal infeasible.
If reasonable conditions could give the specified products in isolable form,
even as a minor component, the proposal is feasible and carries a
\texttt{selectivity\_concern} note. It is infeasible only if no reasonable
conditions could do so. The reason is \texttt{reactivity\_conflict} when a
competing reaction consumes the reactants or the products before the target
can be isolated, and \texttt{regio\_stereo\_violation} when no available
mechanism produces the specified isomer. Because catalysts and conditions may
be assumed, a selectivity that a known catalyst or set of conditions can
achieve counts as obtainable. A stereocenter or substitution pattern that the
transformation does not touch, yet differs between reactants and products, is
not a question of selectivity and is recorded at check~3 as
\texttt{reactants\_cannot\_give\_product}.
Thus, an isolable minor regioisomer remains feasible even when another isomer
predominates.

\input{tab/failure_reasons}

\subsection{Notes on Feasible Proposals}
\label{app:feasible_notes}

Notes document the components or conditions assumed when assigning a feasible label.
The four note categories are:
\begin{description}[leftmargin=1em, itemsep=1pt, topsep=3pt, font=\normalfont\ttfamily]
  \item[no\_identified\_caveat]
    Every species that must be physically added is already listed, and no material concern
    is identified. It does not claim that a complete procedure is given or that the
    experiment would succeed, and it excludes all other notes.
  \item[missing\_auxiliary\_reagent]
    A species permitted by \cref{tab:auxiliary_boundary} is required but not listed, such
    as the base of an alkylation or the reductant of a reductive amination. The annotator
    names it and its role.
  \item[selectivity\_concern]
    Competing chemo-, regio- or stereochemical outcomes are possible, but reasonable
    conditions could still give the products in isolable form.
  \item[harsh\_conditions]
    Credible but demanding conditions, such as elevated temperature or pressure, are
    required. The conditions must remain compatible with the reactants and the products;
    hypothetical extreme conditions cannot rescue an otherwise unsupported
    transformation.
\end{description}
These notes document the assumptions accompanying a feasible label;
the evaluation target is the binary feasibility label.
Difficulty ratings record annotation effort and are not used in the
reported evaluations.

\subsection{Annotators and interface}
\label{app:annotators}

Proposals were labeled by two annotators, a pharmacy faculty member and a postdoctoral
researcher in chemistry, both with extensive knowledge of organic chemistry and
wet-laboratory experience.
Each proposal was presented in a web interface that renders the reactants and the
products as structures from their SMILES, with the SMILES strings available on request.
The annotators had no access to proposal sources or model predictions.
The \goldCalibInPool{} calibration reactions in the pool were settled during
calibration. The \goldKappaN{} reactions outside the calibration set were labeled
independently under the final protocol, without access to each other's labels.

\subsection{Protocol calibration}
\label{app:calibration}

Before the main annotation, both annotators labeled a calibration set of \num{50}
proposals under a draft of the guideline, \goldCalibInPool{} of which belong to the
annotation pool.
The other nine calibration proposals are outside the pool and are not
included in the released benchmark.
Disagreements were discussed by cause, and the discussion was used to amend the
guideline and align the annotators' understanding.
The labels of the calibration proposals were settled in the same discussion, so the
\goldCalibInPool{} that belong to the pool are part of the benchmark but are excluded
from the agreement analysis of \cref{app:agreement}.
The amendments made the substantive--auxiliary test and its structural reading
explicit (\cref{app:criterion}), introduced the ordered decision procedure
(\cref{app:procedure}), and consolidated overlapping failure reasons into the seven of
\cref{tab:failure_reasons}. These amendments were in force for the whole of the main annotation.

\subsection{Agreement and adjudication}
\label{app:agreement}

\Cref{tab:annotator_agreement} gives the independent labels of the two annotators, before
any adjudication, on the \goldKappaN{} reactions of the pool that are not in the
calibration set.
Agreement was \goldAgreePct{}, with Cohen's $\kappa = \goldKappa$
(95\% CI \goldKappaCI). These statistics measure independent judgments
before joint review.

\input{tab/annotator_agreement}

Of the \num{168} disagreements, \num{119} were adjudicated by joint review
against the written guideline. The remaining \num{49} await adjudication and
are excluded from the current benchmark and its reported evaluations.
The benchmark therefore contains \num{591} reactions with matching independent
labels, \num{119} with adjudicated labels, and the \goldCalibInPool{}
calibration reactions, totaling \goldN{} reactions from the \goldPoolN{}-reaction pool.
Recorded reasons and notes are provided as supporting metadata where available.

\subsection{Screening}
\label{app:screens}

Before annotation, every reaction from every source was screened with the general
validity rules applied to the corpus (\cref{tab:filter_rules}).
A reaction was discarded if its SMILES failed to parse or canonicalize, if it contained
an unstable motif such as an \ce{O-O-O} or \ce{B-O-O} chain, if a product appeared
unchanged among its reactants, or if a product contained more heavy atoms of some
element than the reactants.
The screen removes parsing failures and predefined structural patterns.
The remaining reactions still require expert judgment.
No additional structural screening was applied after annotation.

These filters define the structural scope of the benchmark; they do not implement
the substantive--auxiliary distinction. Reactions that pass them are evaluated by
the experts for completeness and reactivity under the stated criterion.

%% file: tab/auxiliary_boundary.tex
\begin{table}[t]
  \centering
  \caption{
    Permitted auxiliary roles. An omitted species may be assumed only when it acts in one
    of these roles, within the stated scope. Any other species is substantive and must be
    listed.
  }
  \label{tab:auxiliary_boundary}
  \small
  \begin{tabular}{@{}>{\raggedright\arraybackslash}p{0.29\textwidth}>{\raggedright\arraybackslash}p{0.65\textwidth}@{}}
    \toprule
    \textbf{Role} & \textbf{Scope} \\
    \midrule
    Catalysis, ligation, solvation
      & Species that supply no retained heavy atom. \\
    \addlinespace[2pt]
    Acid or base assistance
      & Proton transfer or pH control. \\
    \addlinespace[2pt]
    Reduction, hydrogen donation
      & Conventional reducing systems and hydrogen sources that transfer no retained
        heavy atom. \\
    \addlinespace[2pt]
    Oxidation
      & Conventional oxidants that transfer no retained heavy atom. \\
    \addlinespace[2pt]
    Water
      & Water in conventional hydrolysis or workup. \\
    \addlinespace[2pt]
    Activation, dehydration, coupling or deprotection assistance
      & Species that enable these operations without supplying retained heavy atoms,
        within the permitted single-step or one-pot operation. \\
    \bottomrule
  \end{tabular}
\end{table}

%% file: tab/boundary_examples.tex
\begin{table}[t]
  \centering
  \caption{
    The substantive--auxiliary boundary applied to common transformations. Within a row,
    the species in the middle column acts in a permitted role and may be assumed when
    omitted. The species in the right column is a reaction partner and must be listed.
  }
  \label{tab:boundary_examples}
  \small
  \begin{tabular}{@{}lll@{}}
    \toprule
    \textbf{Transformation} & \textbf{Auxiliary (may be assumed)} & \textbf{Substantive (may not)} \\
    \midrule
    Cross-coupling          & Pd catalyst, ligand, base, solvent  & organoboron or halide partner \\
    Esterification          & acid catalyst, activating reagent   & alcohol \\
    Amide coupling          & coupling reagent, base              & amine \\
    Reductive amination     & \ce{NaBH4}, \ce{H2}/Pd              & amine \\
    Wittig olefination      & base                                & ylide or phosphonium salt \\
    Mitsunobu reaction      & DEAD, \ce{PPh3}                     & pronucleophile \\
    Methylation             & base                                & methyl source \\
    Hydrolysis              & water, acid or base                 & --- \\
    \textit{O}-Demethylation & \ce{BBr3}                          & --- \\
    Phthalimide cleavage    & hydrazine                           & --- \\
    Transfer hydrogenation  & isopropanol as hydrogen donor       & --- \\
    \midrule
    \multicolumn{3}{@{}p{0.94\textwidth}@{}}{\emph{Role dependence.} The same species can
      fall on either side: methanol is auxiliary as a solvent and substantive as the source
      of a methyl ester, and isopropanol is auxiliary as a hydrogen donor and substantive as the
      source of an isopropyl ester.} \\
    \bottomrule
  \end{tabular}
\end{table}

%% file: tab/decision_procedure.tex
\begin{table}[t]
  \centering
  \caption{
    Ordered decision procedure. A failure at checks~1--3 ends the procedure;
    otherwise, checks~4--6 are assessed and their reasons may be recorded together.
    Reasons are defined in \cref{tab:failure_reasons}. The two reasons at check~3
    are alternatives (\cref{app:procedure}).
  }
  \label{tab:decision_procedure}
  \small
  \begin{tabular}{@{}c>{\raggedright\arraybackslash}p{0.47\textwidth}>{\raggedright\arraybackslash}p{0.42\textwidth}@{}}
    \toprule
    \textbf{\#} & \textbf{Check} & \textbf{Reason recorded if the check fails} \\
    \midrule
    1 & Can every listed species exist and be present?
      & \texttt{unstable\_or\_nonexistent\_reactant} \\
    \addlinespace[2pt]
    2 & Can the listed products exist as drawn?
      & \texttt{unstable\_or\_nonexistent\_product} \\
    \addlinespace[2pt]
    3 & Can the listed reactants give the products at all?
      & \texttt{missing\_substantive\_reactant} or \newline
        \texttt{reactants\_cannot\_give\_product} \\
    \addlinespace[2pt]
    4 & Can the intended pathway yield isolable products despite competing processes?
      & \texttt{reactivity\_conflict} \\
    \addlinespace[2pt]
    5 & Can the products be obtained with the specified regio- and stereochemistry
        under some reasonable conditions?
      & \texttt{regio\_stereo\_violation} \\
    \addlinespace[2pt]
    6 & Can the products be obtained in one operation, without isolating an intermediate?
      & \texttt{requires\_isolated\_steps} \\
    \midrule
    --- & All checks pass
      & \emph{Feasible}; one or more notes are recorded (\cref{app:feasible_notes}). \\
    \bottomrule
  \end{tabular}
\end{table}

%% file: tab/failure_reasons.tex
\begin{table}[t]
  \centering
  \caption{Failure reasons recorded for infeasible proposals, in the order of the
    checks in \cref{tab:decision_procedure}.}
  \label{tab:failure_reasons}
  \small
  \begin{tabular}{@{}>{\raggedright\arraybackslash}p{0.42\textwidth}>{\raggedright\arraybackslash}p{0.52\textwidth}@{}}
    \toprule
    \textbf{Reason} & \textbf{Definition} \\
    \midrule
    \texttt{unstable\_or\_nonexistent\_reactant}
      & A listed species, substrate or reagent, is drawn incorrectly, cannot exist, or
        is too unstable to be present. Lack of bottle stability alone does not qualify
        a species that can credibly form \emph{in situ}. \\
    \addlinespace[2pt]
    \texttt{unstable\_or\_nonexistent\_product}
      & A listed product has a structural or stability defect that prevents it from
        reasonably existing or being obtained as drawn. Unfamiliarity, or
        failure of software to parse it, does not qualify. \\
    \addlinespace[2pt]
    \texttt{missing\_substantive\_reactant}
      & The intended transformation is recognizable, but a partner that is not a
        permitted auxiliary is not listed. \\
    \addlinespace[2pt]
    \texttt{reactants\_cannot\_give\_product}
      & The listed species can exist, but lack a credible mechanistic route to the products: the
        skeletons do not correspond, a listed reagent cannot perform its required role,
        or there is a concrete mechanistic obstacle. \\
    \addlinespace[2pt]
    \texttt{reactivity\_conflict}
      & A functional group or listed component causes a competing reaction, quenching
        or decomposition that intercepts the intended pathway, so that the products
        cannot be obtained in isolable form within the permitted single-step or one-pot
        operation. \\
    \addlinespace[2pt]
    \texttt{regio\_stereo\_violation}
      & No reasonable conditions give the products with the specified regio- or
        stereochemistry in isolable form. Typical cases are an outcome opposite to
        the one a stereospecific or regiospecific mechanism fixes, and a substitution
        pattern that no available mechanism produces. An isolable minor isomer does
        not qualify. \\
    \addlinespace[2pt]
    \texttt{requires\_isolated\_steps}
      & Obtaining the products requires an additional, independent step with isolation of
        an intermediate. \\
    \bottomrule
  \end{tabular}
\end{table}

%% file: tab/annotator_agreement.tex
\begin{table}[t]
  \centering
  \caption{Independent labels of the two annotators before adjudication, on the
    \num{759} reactions of the annotation pool outside the calibration set. Rows give the first annotator's label and columns the
    second's.}
  \label{tab:annotator_agreement}
  \small
  \begin{tabular}{@{}lrrrr@{}}
    \toprule
    & \multicolumn{3}{c}{\textbf{Annotator 2}} & \\
    \cmidrule(lr){2-4}
    \textbf{Annotator 1} & \textbf{Feasible} & \textbf{Infeasible} & \textbf{Uncertain}
      & \textbf{Total} \\
    \midrule
    Feasible   & \num{255} & \num{77}  & \num{0} & \num{332} \\
    Infeasible & \num{90}  & \num{336} & \num{0} & \num{426} \\
    Uncertain  & \num{0}   & \num{1}   & \num{0} & \num{1}   \\
    \midrule
    Total      & \num{345} & \num{414} & \num{0} & \num{759} \\
    \bottomrule
  \end{tabular}
\end{table}

%% file: appendices/data_curation.tex
\section{Construction of the Benchmark}
\label{app:curation}

\Cref{sec:benchmark} presents the benchmark by source.
This appendix describes how it was built, in the order of construction: curating
positive reactions from ORDerly (\cref{app:positive}), generating negative candidates
(\cref{app:candidates}), collecting the reactions of each source (\cref{app:sources}),
screening and annotation (\cref{app:screen_annotate}), and checking overlap with the
training data (\cref{app:splits}).

\subsection{Positive reactions}
\label{app:positive}

Positive reactions are taken from ORDerly~\citep{wigh2024orderly}, a curated subset of
the Open Reaction Database~\citep{kearnes2021ord} with standardized atom mapping and
yield annotations, starting from its retrosynthesis train and test splits.
All SMILES are canonicalized with RDKit~\citep{rdkit}.
Reactions with a recorded yield of zero are removed from the positives and archived as
candidates for the benchmark (\cref{app:sources}).
We then discard exact duplicates and every reaction that violates a rule marked
\emph{All} or \emph{Positive} in \cref{tab:filter_rules}: reactions that fail parsing,
reactions in which a product component appears verbatim among the reactants, reactions
containing an \ce{O-O-O} or \ce{B-O-O} chain, reactions in which a product contains more
atoms of some heavy element than the reactants, and reactions with three or more highly similar
reactant components.
The curated corpus contains \num{801276} training and \num{76630} test positives.

\subsection{Negative candidates}
\label{app:candidates}

The five generation methods of \cref{subsec:sources} are applied to every positive reaction in
both splits (\cref{app:generation}).
Each candidate inherits the split of the positive reaction from which it was derived and
passes the quality-control pipeline of \cref{app:gen_qc}.
Candidates derived from test-split reactions supply the generated source of the
benchmark; the full output forms the generated corpus analyzed in \cref{sec:negatives}
and \cref{app:corpus_stats}.

\subsection{Collecting the four sources}
\label{app:sources}

\Cref{tab:benchmark_breakdown} breaks the benchmark down by the model, language model
or generation method that produced each reaction.

\paragraph{Retrosynthesis proposals.}
For targets from the test split, we collected reactant proposals from eight
single-step retrosynthesis models: LocalRetro~\citep{chen2021localretro},
G2Retro~\citep{chen2023g2retro}, UAlign~\citep{zeng2024ualign},
EditRetro~\citep{han2024editretro}, Chemformer~\citep{irwin2022chemformer},
LlaSMol~\citep{yu2024llasmol}, and LARC~\citep{baker2025larc} with two backbone language
models, Claude 3.5 Sonnet~\citep{anthropic2024claude35sonnet} and Mistral
NeMo~\citep{mistral2024nemo}.
Proposals were pooled across models and deduplicated by canonical SMILES.
They were retained whether or not they reproduce the recorded reactants, since both kinds
belong to the distribution a model faces in deployment.

\paragraph{Zero-yield records.}
From ORDerly we retrieve the reactions with a recorded yield of exactly zero that were
removed during positive curation, apply the same canonicalization and validity rules, and
remove any reaction whose canonical reactant--product pair also occurs among the
positives.
A zero-yield record shows only that the conditions tried did not give the product
(\cref{app:rationale}); these reactions are labeled under the same criterion as every
other source.

\paragraph{LLM perturbations.}
Recorded reactions from the test split were given to GPT-5.4~\citep{openai2026gpt54} and
Claude Opus 4.6~\citep{anthropic2026claude46}, which were instructed
to modify the reactants or the product so that the reaction becomes infeasible while
remaining superficially coherent.
The instruction states the models' intent only; whether a modified reaction is infeasible
is decided by the annotators.

\paragraph{Generated reactions.}
Generated candidates were sampled in equal numbers from the test-split output of each of
the five generation methods (\cref{app:generation}).
All \num{289} are outputs of these methods, including those the annotators judged feasible.

\input{tab/benchmark_breakdown}

\subsection{Screening and annotation}
\label{app:screen_annotate}

Before annotation, every reaction from every source passed the screen described in
\cref{app:screens}.
The screened pool of \goldPoolN{} reactions was then labeled under the protocol of
\cref{app:annotation}. The calibration, independent annotation and adjudication
procedure yields the \goldN{} benchmark reactions (\cref{app:agreement}).

\subsection{Splits and overlap}
\label{app:splits}

The corpus follows the ORDerly retrosynthesis train/test split, and all benchmark
reactions except the zero-yield records derive from the test split.
No benchmark reactant--product pair occurs in the data used to train our models.
Product overlap remains possible even when the complete reaction differs.
\num{115} benchmark reactions have a main product that is also the product of a training
reaction, \num{97} of them retrosynthesis proposals.
BARTSmiles log-likelihood scores \num{76.6} AUROC on those \num{97} proposals
against \num{80.8} on the other \num{185}; this comparison is descriptive because
the two subsets may differ in difficulty.

%% file: tab/benchmark_breakdown.tex
\begin{table}[t]
  \centering
  \caption{Breakdown of the benchmark by the model, language model or negative candidate
    generation method that produced each reaction. \emph{Infeas.\ (\%)} is the share of a
    row's reactions the experts judged infeasible, which for a generation method is the
    share of its candidates that are genuinely negative.}
  \label{tab:benchmark_breakdown}
  \small
  \begin{tabular}{@{}llrrrr@{}}
    \toprule
    \textbf{Source} & \textbf{Origin} & \textbf{Feasible} & \textbf{Infeasible} & \textbf{Total} & \textbf{Infeas.\ (\%)} \\
    \midrule
    RP & LARC (Claude 3.5 Sonnet) & \num{43} & \num{12} & \num{55} & 21.8 \\
       & LARC (Mistral NeMo) & \num{42} & \num{12} & \num{54} & 22.2 \\
       & G2Retro      & \num{17} & \num{13} & \num{30} & 43.3 \\
       & EditRetro    & \num{18} & \num{12} & \num{30} & 40.0 \\
       & LlaSMol      & \num{16} & \num{13} & \num{29} & 44.8 \\
       & UAlign       & \num{14} & \num{15} & \num{29} & 51.7 \\
       & Chemformer   & \num{22} & \num{6}  & \num{28} & 21.4 \\
       & LocalRetro   & \num{25} & \num{2}  & \num{27} & 7.4 \\
    \midrule
    ZR & ORDerly zero-yield records & \num{79} & \num{14} & \num{93} & 15.1 \\
    \midrule
    LP & GPT-5.4      & \num{15} & \num{30} & \num{45} & 66.7 \\
       & Claude Opus 4.6 & \num{17} & \num{25} & \num{42} & 59.5 \\
    \midrule
    GR & \negraabbr   & \num{8}  & \num{50} & \num{58} & 86.2 \\
       & \negrrabbr   & \num{4}  & \num{53} & \num{57} & 93.0 \\
       & \negncabbr   & \num{4}  & \num{54} & \num{58} & 93.1 \\
       & \negftabbr   & \num{13} & \num{44} & \num{57} & 77.2 \\
       & \negetabbr   & \num{4}  & \num{55} & \num{59} & 93.2 \\
    \midrule
    \multicolumn{2}{@{}l}{\textbf{Total}} & \num{341} & \num{410} & \num{751} & 54.6 \\
    \bottomrule
  \end{tabular}
\end{table}

%% file: appendices/negative_generation_details.tex
\section{Negative Candidate Generation}
\label{app:generation}

The generated reactions in the benchmark and the training corpus are produced by five
generation methods.
Each produces candidates intended to be infeasible, and whether a candidate is in fact
infeasible is established only by expert annotation (\cref{subsec:annotation_quality}).
This appendix summarizes the procedures, their main settings, and the quality-control
pipeline.

\subsection{Design}
\label{app:generation_design}

Each method perturbs a positive reaction $(\mathbf{R}^*, \mathbf{P}^*)$ along one of
two axes.
Reactant perturbations (\negraabbr, \negrrabbr, \negncabbr) replace $\mathbf{R}^*$ with
$\tilde{\mathbf{R}}$ while holding the products fixed, whereas outcome perturbations
(\negftabbr, \negetabbr) replace $\mathbf{P}^*$ with $\tilde{\mathbf{P}}$ while holding
the reactants fixed.
All five methods are applied to every positive reaction in both splits, and each
candidate retains a pointer to its parent positive, which supports pairwise training and
method-level analysis.

\subsection{Random Addition (\negraabbr)}
\label{app:gen_ra}

\negraabbr modifies one or more reactants at the reaction center.
Each reactant is reduced to its synthon by removing leaving groups, atoms drawn from a
frequency-weighted distribution of common leaving-group atoms (e.g., \ce{C}, \ce{O},
\ce{N}, halogens) are appended iteratively at the reaction center under valence
constraints, for at most six addition steps per synthon, and the modified synthons are
reassembled into reactants $\tilde{\mathbf{R}}$, giving the candidate
$(\tilde{\mathbf{R}}, \mathbf{P}^*)$. Up to three candidates are generated per positive
reaction.
The modification is intended to make a reactant incompatible with the required
bond-forming event.
It does not guarantee a difficult candidate, and it can produce a reactant that still
undergoes a related transformation.

\subsection{Random Replacement (\negrrabbr)}
\label{app:gen_rr}

\negrrabbr replaces the reactants with structurally similar reactants drawn from the
positive training set.
For each positive reaction, we retrieve the 200 most similar reactions by Tanimoto
similarity on a joint reactant--product fingerprint and take as $\tilde{\mathbf{R}}$ the
reactants of the highest-ranked ones that are similar to, but different from,
$\mathbf{R}^*$. Up to nine candidates are formed per positive reaction, eight in this way
and one from a reactant set drawn at random from the training set.
The procedure does not verify that the replacement fails to give $\mathbf{P}^*$, and it
discards replacements whose Tanimoto similarity to the original reactant falls below $0.1$
(\cref{tab:filter_rules}).

\subsection{New Center (\negncabbr)}
\label{app:gen_nc}

\negncabbr proposes an alternative disconnection of the product.
The reaction-center model of G2Retro~\citep{chen2023g2retro} predicts the top-10
candidate centers for $\mathbf{P}^*$, and up to three centers that differ from the
recorded one are retained, taken from the lowest-ranked end of the list.
The resulting synthons are completed by RLSynC~\citep{baker2024rlsync}, which returns ten
completions ranked by its value estimate, and the low-ranked completions are selected.
A low rank indicates only that a completion is unlikely under RLSynC, not that the
resulting proposal is infeasible.

\subsection{Forward Template (\negftabbr)}
\label{app:gen_ft}

\negftabbr applies forward reaction templates extracted from the positive training set
to the recorded reactants, using RDChiral~\citep{coley2019rdchiral}.
We apply every template that occurs at least ten times in the training set together with
up to ten randomly drawn rarer templates per reaction, and every product $\tilde{\mathbf{P}}$ that differs from $\mathbf{P}^*$ under canonical
SMILES comparison gives a candidate $(\mathbf{R}^*, \tilde{\mathbf{P}})$.
A product predicted by a template but not recorded for these reactants may nevertheless
be feasible.

\subsection{Elementary Template (\negetabbr)}
\label{app:gen_et}

\negetabbr enumerates elementary steps with 739 templates from the library of elementary
templates of~\citet{joung2024elementary}.
Starting from $\mathbf{R}^*$, the templates are applied breadth-first for up to four
steps, and every product reached in at least two steps that differs from $\mathbf{P}^*$
and is not recorded as a product of these reactants gives a candidate, with at most 50
candidates per reaction.
As for \negftabbr, an unrecorded product is not necessarily infeasible.

\subsection{Quality control}
\label{app:gen_qc}

All candidates pass a unified pipeline of 16 rules in three tiers: general validity,
negative--positive consistency, and removal of method-specific artifacts
(\cref{tab:filter_rules}).
In particular, no candidate may encode a product set that is a recorded outcome of the
same reactants in the positive corpus.

\input{tab/filter_rules}

\subsection{Size of the generated corpus}
\label{app:corpus_stats}

\Cref{tab:dataset_stats} gives the size of the generated corpus by split and method.
Per-method rows count candidates before cross-method deduplication, whereas the
\emph{Negative} row counts unique canonical reactions. The per-method rows therefore
sum to \num{20772} (train) and \num{1408} (test) more than the \emph{Negative} row.
These excess counts arise from candidates produced by more than one method.

\input{tab/dataset_stats}

%% file: tab/filter_rules.tex
\begin{table}[ht]
\centering
\caption{%
  Summary of the 16 filtering rules used to construct the generated corpus.
  ``All'' denotes filters applied to both recorded positives and generated candidates;
  these general filters also screen the benchmark before annotation.
  Methods are described in \cref{app:generation}.
}
\label{tab:filter_rules}
\small
\setlength{\tabcolsep}{5pt}
\begin{tabular}{llp{6.5cm}}
\toprule
\textbf{Rule} & \textbf{Scope} & \textbf{Condition for invalidation} \\
\midrule
\multicolumn{3}{l}{\textit{General validity}} \\
\texttt{all\_invalid\_smiles}            & All       & Canonical reactant or product SMILES is null \\
\texttt{product\_in\_reactant}           & All       & Any product component appears verbatim among the reactant components \\
\texttt{unstable\_substructure}          & All       & Reactant or product contains an \ce{O-O-O} or \ce{B-O-O} chain \\
\texttt{atom\_conservation}              & All       & Products contain more atoms of any heavy element than the reactants \\
\texttt{reactant\_internal\_similarity}  & Positive  & Median pairwise reactant Tanimoto $> 0.5$ with $\geq\!3$ components \\
\midrule
\multicolumn{3}{l}{\textit{Negative--positive consistency}} \\
\texttt{negative\_parent\_not\_none}     & Negative  & \texttt{parent} field is null \\
\texttt{negative\_split\_match\_parent}  & Negative  & Negative split $\neq$ parent split, or parent not found \\
\texttt{negative\_parent\_not\_invalid}  & Negative  & Parent positive has been marked invalid \\
\texttt{ra\_nc\_reaction\_centers}       & \negraabbr, \negncabbr & Parent has null \texttt{reaction\_centers} annotation \\
\texttt{neg\_is\_not\_pos}               & Negative  & Product set $\subseteq$ some positive's product set for the same reactant SMILES \\
\midrule
\multicolumn{3}{l}{\textit{Method-specific artifacts}} \\
\texttt{ft\_small\_product}              & \negftabbr        & Product has $<\!6$ heavy atoms \\
\texttt{ft\_stereo\_only}                & \negftabbr        & Product equals parent product after stripping stereochemical annotations \\
\texttt{ft\_et\_tautomer\_equiv}         & \negftabbr, \negetabbr & Product is a strict tautomer of the parent product \\
\texttt{rr\_reactant\_too\_dissimilar}   & \negrrabbr        & Tanimoto(replacement reactant, original reactant) $< 0.1$ \\
\texttt{nc\_small\_fragment}             & \negncabbr        & Reactant set contains a blacklisted, chemically unreasonable fragment \\
\texttt{nc\_spurious\_ci}                & \negncabbr        & Methyl iodide present in reactants but contributes no net methylation to the product \\
\bottomrule
\end{tabular}
\end{table}

%% file: tab/dataset_stats.tex
\begin{table}[t]
  \centering
  \caption{Size of the generated corpus by split. Per-method rows count candidates
    before cross-method deduplication; the \emph{Negative} row counts unique canonical
    reactions.}
  \label{tab:dataset_stats}
  \small
  \begin{tabular}{@{}lrr@{}}
    \toprule
    & \textbf{Train} & \textbf{Test} \\
    \midrule
    Positive & \num{801276} & \num{76630} \\
    Negative (unique) & \num{12047157} & \num{1189994} \\
    \quad \negraabbr & \num{1705924} & \num{163253} \\
    \quad \negrrabbr & \num{3024969} & \num{295737} \\
    \quad \negncabbr & \num{1433340} & \num{149175} \\
    \quad \negftabbr & \num{1781362} & \num{181411} \\
    \quad \negetabbr & \num{4122334} & \num{401826} \\
    \midrule
    \textbf{Total (unique)} & \num{12848433} & \num{1266624} \\
    \bottomrule
  \end{tabular}
\end{table}

%% file: appendices/system_implementations.tex
\section{System Implementations}
\label{app:systems}

This appendix describes the evaluated systems and their training settings.
Model selection, decision thresholds and statistical tests are given in
\cref{app:protocol}.

\input{appendices/evaluation_details.tex}

\input{appendices/verifier_model_details.tex}

%% file: appendices/evaluation_details.tex
\subsection{LLMs and Baselines}
\label{app:baselines}

\subsubsection{Large Language Models}
\label{app:llm_baseline}

We query Claude Opus 5 and GPT-6 Astra through their APIs with extended reasoning at the
\texttt{high} setting, one call per reaction, with no examples from \dataname and no
parameter updates. Both models receive the same
system prompt, verified identical byte for byte. It states the criterion of
\cref{sec:task} operationally: what may
be assumed without being listed, what may not, and six conditions under which the reaction
is not feasible, which correspond to the failure reasons of \cref{app:reasons} with the
third condition covering two of them. The user message
carries the reactant and product SMILES.

The decision is read from the final answer tag. One Claude Opus 5 response contained no
parsable tag and is excluded from its metrics. Claude Opus 5
used on average \num{1073} prompt and \num{532} completion tokens per reaction, at a total
cost of US\$30.31 for the \goldN{} reactions (US\$0.040 per reaction).
GPT-6 Astra used on average \num{709} prompt
and \num{1032} completion tokens, \num{956} of them for reasoning, at a total cost of
US\$44.07 (US\$0.059 per reaction) and 32.6 seconds per call.

\subsubsection{Forward-Model Proxies}
\label{app:forward_baseline}

Chemformer~\citep{irwin2022chemformer} and BARTSmiles~\citep{chilingaryan2024bartsmiles} are
fine-tuned for forward prediction on the recorded reactions of the training split, so
neither sees a negative candidate. Chemformer starts from its released checkpoint and is
trained for 20 epochs. BARTSmiles starts from the Stage-1 checkpoint of \cref{app:model} and
is trained for \num{15490} steps. Each is taken at the checkpoint with
the lowest validation loss: epoch 17 for Chemformer (validation loss 0.04685, against
0.04687 at the last epoch) and step \num{12000} for BARTSmiles (0.04678, against 0.04811 at
step \num{15000}). The validation reactions of BARTSmiles largely overlap with those used in
Stage 1, so its validation loss is optimistic in absolute terms. It serves only to pick a
checkpoint.

For the top-10 readout, each model generates 50 beams, and a reaction is accepted if the
canonical SMILES of its products matches one of the ten highest-ranked predictions. The
log-likelihood readout is the mean per-token log-probability of the product string given
the reactant string under teacher forcing, computed with a maximum length of \num{1700}
tokens for BARTSmiles and the default length of Chemformer.

\subsubsection{The AiZynthFinder In-Scope Filter}
\label{app:aizynth}

\paragraph{Evaluation setup.}
We evaluate the released AiZynthFinder filter as an existing planning baseline.
The model embeds ECFP4 fingerprints of the product and the difference between
product and reactant fingerprints, and scores them by cosine similarity.
It has about 4.2M parameters and was trained on \num{911869} USPTO reactions
and \num{1621484} template-generated negatives~\citep{genheden2020filter}.
We use \texttt{uspto\_filter\_model.onnx} from version 4.4.1 without retraining,
through AiZynthFinder's own reaction objects and filter implementation.
No reactant failed to parse. We report results using both the threshold
rule of \cref{app:thresholds} and the filter's default cutoff of 0.05;
AUROC is independent of the cutoff.

\paragraph{Potential training-data overlap.}
The training set is not released. We check against Lowe's raw USPTO
extraction~\citep{lowe2012extraction}, a superset of the training data,
to bound possible overlap. Matching up to reagents, \num{132} benchmark
reactions appear there: 92 of 93 ZR, 33 of 87 LP, 7 of 282 RP and none
of the 289 GR reactions. These matches indicate possible exposure,
particularly for ZR and LP, rather than verified membership in the
filter's training set. On the \num{619} reactions with no match,
its pooled AUROC is 63.3.

\paragraph{Performance across generation methods.}
On the held-out generated pairs of \cref{app:thresholds}, AUROC is 89.0
for \negftabbr, 83.6 for \negetabbr and 58.0 for \negncabbr.
Three additional released filters trained on individual negative types
show the same ordering.

\paragraph{Scope of the comparison.}
These results characterize how the released filter transfers to \dataname.
Its training data and input conventions differ from ours: its inputs omit
reagents, its fingerprints ignore stereochemistry, and its negative labels
come from template-generated candidates. Removing carbon-free fragments
from the benchmark's reactant lists changes pooled AUROC from 65.9 to 65.3.
The benchmark comparison therefore evaluates the existing tool under our
criterion. \Cref{sec:negatives} explores changes to negative training data
using matched training runs.

\subsubsection{Feature-Based Classifier}
\label{app:shallow}

\paragraph{Features.}
Twenty-two additive descriptors are computed for the products and as the difference
between products and reactants: molecular weight, heavy atoms, rings, aromatic rings,
rotatable bonds, hydrogen-bond donors and acceptors, topological polar surface area, logP,
formal charge, bonds, stereocenters, and the counts of C, N, O, S, P, F, Cl, Br, I, and
other heavy elements. To these 44 features we add the numbers of reactants and products,
the product-to-reactant ratios of molecular weight and heavy atoms, the lengths of both
SMILES strings and their ratio, the excess of heavy atoms in the products and the number of
elements whose count increases, the maximum, mean, and minimum Tanimoto similarity between
the Morgan fingerprints (radius 2, \num{2048} bits) of the largest product and each
reactant, and two indicators of whether the product is a substructure of a reactant or the
reverse, for 58 features in total, all computed with RDKit~\citep{rdkit}.

\paragraph{Model and data.}
We train a histogram gradient-boosted tree classifier~\citep{pedregosa2011scikit} with 400
iterations, learning rate 0.06, at most 31 leaves, at least 20 samples per leaf, L2
regularization 1.0, and early stopping on an internal 10\% split. The training data are
\num{100000} pairs from the training split, \num{20000} per method, in which every recorded
reaction appears at most once. None of the \num{200000} rows coincides with a benchmark
reaction, even after sorting reactant fragments. The predicted probability of feasibility is
the score, thresholded at \num{0.510}, the value that maximizes BAcc on the held-out generated
pairs of \cref{app:thresholds}.

\paragraph{Benchmark performance.}
\gbdtname reaches 70.7 mean AUROC on \dataname. On the per-method GR
comparison, it reaches 87.0 on NC but is near chance on RA and RR
(\cref{tab:results_per_method}). The descriptors therefore provide a
useful reference, with performance varying substantially by candidate type.

\paragraph{Separability of each method's candidates.}
On the held-out 10\% of the training pairs, about \num{2000} candidates of each method are
scored against \num{10000} recorded reactions (\cref{tab:app_shallow_sep}). Before running
this analysis we predicted that \negrrabbr candidates, which use unmodified reactants
from other recorded reactions, would be the hardest to separate. The prediction holds
among the three methods with similar estimated negative purities. \negraabbr and
\negrrabbr have similar separability here, which may reflect \negraabbr's lower
estimated purity or the limited descriptor changes caused by its edits.

\begin{table}[h]
  \centering
  \caption{AUROC of \gbdtname in separating each method's negative candidates
  from recorded reactions on held-out training pairs, with DeLong 95\% confidence intervals.
  Negative purity is estimated separately from the expert-labeled GR subset
  (\cref{tab:benchmark_breakdown}).}
  \label{tab:app_shallow_sep}
  \small
  \begin{tabular}{@{}l c r c@{}}
    \toprule
    Method & Purity (\%) & Candidates & AUROC \\
    \midrule
    \negncabbr & 93.1 & \num{1981} & 98.5 {\scriptsize\color{gray}[98.2, 98.8]} \\
    \negftabbr & 77.2 & \num{1970} & 92.5 {\scriptsize\color{gray}[91.7, 93.2]} \\
    \negetabbr & 93.2 & \num{1985} & 91.5 {\scriptsize\color{gray}[90.8, 92.2]} \\
    \negrrabbr & 93.0 & \num{2002} & 86.4 {\scriptsize\color{gray}[85.6, 87.2]} \\
    \negraabbr & 86.2 & \num{2062} & 85.9 {\scriptsize\color{gray}[85.1, 86.7]} \\
    \bottomrule
  \end{tabular}
\end{table}

\paragraph{What the model uses.}
Permutation importance on the benchmark ranks the change in bond count, the change in
carbon count, and the two substructure indicators highest. The heavy-atom excess and the
number of elements whose count increases have zero importance, consistent with the
pre-annotation screen removing these element-count excesses (\cref{app:screens}).
Fitting the same model directly on the benchmark by five-fold cross-validation gives
a pooled AUROC of 75.0, compared with 79.8 for the model trained on generated pairs.
These scores reflect both reaction-level discrimination and source composition
(\cref{subsec:composition}), so they do not by themselves establish how well these
descriptors capture feasibility within a source.

\subsubsection{Rule-Based Baseline}
\label{app:rule_baseline}

\rulemethodname is a non-learning cascade implemented with RDKit. It rejects a reaction if
(1)~a product cannot be parsed, (2)~a molecule contains an unstable motif such as an
\ce{O-O-O} or \ce{B-O-O} chain, (3)~the products contain more atoms of some heavy element
than the reactants, (4)~a product appears unchanged among the reactants, (5)~no reactant has a
Morgan-fingerprint similarity to the product above a small threshold, or (6)~a functional
group formed in the product, such as an amide, ester, ether, imine, urea, carbamate,
sulfonamide, nitrile, or nitro group, has no compatible precursor among the reactants,
unless the product is a substructure of a reactant or is highly similar to one. Rules (1)
to (4) coincide with the pre-annotation screen, so on \dataname only rules (5) and (6) can
reject a reaction. The baseline averages 60.2 BAcc over the four sources
(\cref{tab:app_full_auroc}), and we omit it from the main text.

%% file: appendices/verifier_model_details.tex
\subsection{Models Trained with Generated Negative Candidates}
\label{app:model}

This appendix describes \discname, the six other discriminative verifiers
reported in \cref{tab:app_family_matrix}, and the controlled comparisons of
\cref{sec:negatives}.
All models use random seed 42. The settings held fixed within each experiment
family are specified below.

\subsubsection{The Discriminative Verifier}
\label{app:model_main}

\paragraph{Backbone and input.}
\discname is built on BARTSmiles~\citep{chilingaryan2024bartsmiles}, a BART-style
encoder--decoder model pre-trained on SMILES strings. A reaction is written as
\texttt{\{reactants\}[[>>]]\{products\}[[Judge]]}, where molecules are canonicalized with
RDKit~\citep{rdkit} and joined by periods. There are no separate reagent or condition
fields; any species listed among the reactants remains in the input. The
tokens \texttt{[[>>]]}, \texttt{[[Judge]]}, and \texttt{[[Yes]]} are added to the vocabulary
with randomly initialized embeddings. The whole string is the encoder input.
The decoder receives its start token, and the logit of \texttt{[[Yes]]} at the first
decoding position is the feasibility score $s(\mathbf{R}, \mathbf{P}) \in \mathbb{R}$.
Scoring a reaction therefore costs one encoder pass and one decoder step.

\paragraph{Stage 1: reaction denoising.}
Reaction strings are longer than the single molecules seen in pre-training, so the model
is first adapted to them by denoising. On the recorded reactions of the training split,
contiguous token spans with lengths drawn from a Poisson distribution ($\lambda = 3.5$) are
corrupted, up to 20\% of the non-special tokens. Corrupted positions are replaced by a
\texttt{<mask>} token or by random tokens, and the decoder reconstructs the original string under the usual cross-entropy
loss. The checkpoint with the lowest validation reconstruction loss initializes Stage 2.

\paragraph{Stage 2: pairwise training.}
Each negative candidate in the training split is paired with the recorded reaction from
which it was generated. Candidates that coincide after canonicalization are merged, and a
candidate produced by several methods is assigned to one of them so that all five methods
contribute equally. All parent reactions in the resulting \num{7095035} pairs belong
to the positive training corpus of \cref{app:positive}. For a batch of $B$ pairs with
scores $s_i^+$ (recorded reaction)
and $s_i^-$ (negative candidate), the loss is
\begin{equation}
  \mathcal{L} = \alpha_{\mathrm{BT}}\,\mathcal{L}_{\mathrm{BT}}
              + \alpha_{\mathrm{BCE}}\,\mathcal{L}_{\mathrm{BCE}}
              + \alpha_{\mathrm{C}}\,\mathcal{L}_{\mathrm{C}},
  \qquad (\alpha_{\mathrm{BT}}, \alpha_{\mathrm{BCE}}, \alpha_{\mathrm{C}}) = (0.8, 0.1, 0.1),
\end{equation}
with
\begin{align}
  \mathcal{L}_{\mathrm{BT}}  &= -\frac{1}{B}\sum_{i=1}^{B} \log \sigma\!\left(s_i^+ - s_i^-\right), \\
  \mathcal{L}_{\mathrm{BCE}} &= -\frac{1}{2B}\sum_{i=1}^{B}
      \left[\log \sigma\!\left(s_i^+\right) + \log\!\left(1 - \sigma\!\left(s_i^-\right)\right)\right], \\
  \mathcal{L}_{\mathrm{C}}   &= \left(\frac{1}{2B}\sum_{i=1}^{B}\left(s_i^+ + s_i^-\right)\right)^{2},
\end{align}
where $\sigma$ is the logistic function. The Bradley--Terry term $\mathcal{L}_{\mathrm{BT}}$
ranks each recorded reaction above its candidate and is unchanged when a constant is added
to all scores. The cross-entropy term removes this invariance by tying the scores to a
decision boundary at zero, and the centering term keeps the mean score of a batch near
zero. The cross-entropy term alone would not bound the scores, since it keeps decreasing as
they grow apart. The centering term controls the batch mean, while weight decay
regularizes the model parameters.

\paragraph{Dynamic re-weighting.}
Methods differ in how quickly the model learns their candidates. Every $T$ steps, the
pairwise accuracy $\mathrm{acc}_m$ of the current model on held-out pairs of method $m$ is
measured, and pairs of method $m$ are then sampled with probability
\begin{equation}
  w_m \propto \exp\!\left(\frac{1 - \mathrm{acc}_m}{\tau}\right),
\end{equation}
where $\tau$ is a temperature, and no method falls below a sampling probability of 0.05.
\discname uses $T = \num{5000}$ steps.

\paragraph{Decision rule.}
A reaction is predicted feasible if $s(\mathbf{R}, \mathbf{P}) > t$, where $t$ is the raw
logit threshold frozen on held-out generated pairs (\cref{tab:app_thresholds}).

\subsubsection{The Discriminative Verifier Family}
\label{app:family}

\Cref{tab:app_family} lists the seven discriminative verifiers. All
start from the same Stage-1 checkpoint and use the same Stage-2 loss, optimizer settings,
and effective batch of 512 pairs. They are compared at
step \num{20000}, where each has seen \num{10.24} million pairs. The number of distinct
recorded reactions in the training pairs differs across methods. \negftabbr has the
fewest, each appearing 8.6 times on average.

\begin{table}[h]
  \centering
  \caption{Training configuration of the seven discriminative verifiers. All are evaluated at step
  \num{20000}. A balanced pool contains equal numbers of candidates from each method;
  dynamic re-weighting changes their sampling probabilities during training.}
  \label{tab:app_family}
  \small
  \begin{tabular}{@{}l r c r@{}}
    \toprule
    Negative candidates & Pairs & Re-weighting & Steps trained \\
    \midrule
    One method (\negraabbr, \negrrabbr, \negncabbr, \negftabbr, or \negetabbr) & \num{1419007} & none & \num{22168} \\
    All five methods, balanced pool & \num{1419007} & every \num{1000} steps & \num{22168} \\
    All five methods (\discname) & \num{7095035} & every \num{5000} steps & \num{41571} \\
    \bottomrule
  \end{tabular}
\end{table}

\subsubsection{The Controlled Pairs}
\label{app:controlled_pair}

The models of \cref{sec:negatives} start from the BARTSmiles forward model of
\cref{app:forward_baseline} and are trained for one epoch, with identical order, seed,
learning-rate schedule, warmup, and numerical precision within a pair. The smaller pair uses
\num{800000} pairs (\num{6250} steps) and the larger \num{2400000} (\num{18750} steps), at
an effective batch of 128. The first optimizes the forward-prediction loss
$-\log p_\theta(\mathbf{P} \mid \mathbf{R})$ on the recorded reaction of each pair, and its
negative candidates are loaded but unused. The second adds a weighted Bradley--Terry term
like $\mathcal{L}_{\mathrm{BT}}$, computed on the score used at inference, the mean
per-token log-likelihood $\bar\ell_\theta(\mathbf{R}, \mathbf{P})$, of the recorded
reaction and of its candidate. Every run is evaluated at the end of its epoch, so no checkpoint is selected. Selecting by
validation loss would break the pairing, because the loss of the run with negatives contains
the preference term and the loss of the run without it does not, so the two are not the same
quantity. Benchmark AUROC varies by at most 1.0 point over the second half of each
run. All four are thresholded like every other continuous score.

\subsubsection{The Per-Method Comparison, and Its Pre-Registration}
\label{app:prereg}

The seven groups of \cref{tab:neg_permethod} share the recipe of \cref{app:controlled_pair}
at the \num{800000}-pair scale and differ only in the pairs the preference term sees: none
(the reference), \num{800000} from one method, or \num{160000} from each of the five. The
forward-prediction loss is computed on one stream of recorded reactions, in one order,
shared by all seven. This holds the forward-prediction signal fixed while varying
the pairs supplied to the preference loss. In the discriminative family, positive
examples are drawn from each method's own pairs, so the method comparisons also
vary positive coverage. The mixture uses plain concatenation with no re-weighting.
Every group is trained for one epoch of \num{6250} steps from the same initialization with
identical optimizer settings and seed, and is scored at the final step.

The design, the endpoints and the analysis were written down before any training data was
sampled and before any run was launched. The
primary endpoint is each group's four-source mean AUROC minus the reference group's, tested
by a paired bootstrap stratified by source with Holm correction across the six groups. The
registration also fixed the rule that conclusions about methods are drawn from the three
with similar estimated negative purities, \negrrabbr, \negncabbr and \negetabbr, with
\negftabbr and \negraabbr reported for reference. It recorded three predictions: that
\negrrabbr would not be lower than any other group receiving negatives, that \negftabbr
would be the lowest of the five single-method groups, and that the mixture would not exceed
the reference on RP. The observed point estimates follow all three predictions.

One deviation is recorded. The registration required the sampled training pairs to contain
no benchmark reaction and the experiment to stop otherwise. The first sample contained
four such pairs, the runs were stopped, and each was replaced by the next unused clean pair
of the same permutation before the runs were restarted. No stream shares a reaction with
the benchmark.

The positive-only control is trained separately for this experiment using its shared
positive stream. Its mean AUROC is 71.7, compared with 71.4 for the smaller control
in \cref{tab:neg_controlled}; the respective ZR scores are 79.0 and 76.5.
Each negative-supervision effect is therefore measured against the control trained
under the corresponding comparison design.

%% file: appendices/protocol_details.tex
\section{Evaluation Protocol Details}
\label{app:protocol}

\subsection{Metrics}
\label{app:metrics}

AUROC is the probability that a randomly drawn feasible reaction receives a higher score
than a randomly drawn infeasible one, with ties counted as one half. We compute it within a
source. TPR is the share of feasible reactions a system accepts, TNR the share of
infeasible reactions it rejects, and BAcc their mean. Within a fixed source, BAcc
weights the two classes equally. The average of a system is the unweighted mean
over the four sources and was fixed before the evaluation. It includes GR, whose
generation methods also supply the negative training data for several systems.
The AUROC of a negative candidate
method on GR compares that method's infeasible candidates with all 33 feasible GR
candidates, which come from all five methods (\negraabbr 8, \negrrabbr 4, \negncabbr 4,
\negftabbr 13, \negetabbr 4).

\paragraph{Why we evaluate within sources.}
\label{subsec:composition}
Feasible rates range from 11.4\% in GR to 84.9\% in ZR. To illustrate the
effect on a pooled score, each reaction
receives the fraction of feasible labels in its source in
\cref{tab:benchmark_composition}. We then compute AUROC over all reactions.
Every within-source comparison is tied and contributes one half, but the
between-source comparisons raise pooled AUROC to 82.1. These source rates
are estimated from the benchmark labels; the calculation illustrates what
pooled AUROC can reward without assessing individual reactions.
Pooling also reverses a model comparison: \discname exceeds Chemformer LL
in pooled AUROC (82.5 versus 81.9), but is lower in four-source mean
AUROC (69.7 versus 72.6). Evaluating within each source prevents
between-source comparisons from dominating the reported performance.

\subsection{Sensitivity to Auxiliary-Reagent Annotations}
\label{app:auxsens}

We assess the effect of auxiliary-reagent annotations by excluding the
\num{104} feasible reactions carrying \texttt{missing\_auxiliary\_reagent}
and evaluating all systems at the same frozen thresholds
(\cref{tab:app_auxsens}). This subset is defined from annotation metadata
without reference to model predictions. The note records a permitted
auxiliary omission under \cref{app:criterion}.

Claude Opus 5 retains the highest mean BAcc, while some intermediate
rankings change. The four leading mean AUROCs range from 74.6 to 75.0
on the restricted subset: \negname is comparable to the forward
baselines here, rather than retaining its full-benchmark lead.
Its advantage over \discname remains 4.2 points.

\input{tab/app_auxsens}

\subsection{Checkpoints}
\label{app:checkpoints}

Forward models are taken at the checkpoint with the lowest validation loss for forward
prediction (\cref{app:forward_baseline}). For the seven discriminative verifiers, we scored all 129
saved checkpoints on the \num{10000} held-out generated pairs of \cref{app:thresholds}. For
every model, between 38.9\% and 75.0\% of the checkpoints after step \num{5000} lie within
the DeLong 95\% confidence interval of the best checkpoint, and for \discname the best and
the final checkpoints differ by only 0.002 AUROC points. In-domain validation
provides little separation among these checkpoints. We report a common budget
of \num{20000} steps, the latest
step at which all seven runs saved a checkpoint. At this step every model has seen
\num{10.24} million pairs, whereas the final checkpoints of the runs differ by a factor of
1.9 in compute. For \discname, step \num{20000} also gives the highest pooled AUROC on RP,
ZR, and LP among its nine checkpoints, which could favor it.

\Cref{tab:app_checkpoints} compares the reported checkpoints with the final ones and with
an oracle chosen on the benchmark labels. \discname stays at or below 67.8 AUROC on RP
at every checkpoint, and even its oracle average (71.2) falls short of forward-model
likelihood (72.6 and 72.8). The verifier trained on \negrrabbr candidates remains the best
of the seven under all three rules, the one trained on \negftabbr candidates remains near
chance, and the small mixture stays close to the full one.

\subsection{Thresholds}
\label{app:thresholds}

Every continuous score is thresholded at the value that maximizes BAcc on \num{10000}
held-out pairs of recorded reactions and negative candidates from the generated test split.
The set is balanced between recorded reactions and candidates and across the five methods,
and shares no reactant or product string with the benchmark. BAcc gives equal
weight to the two classes, but the selected threshold can still depend on the
mixture of generation methods. The threshold is then frozen
(\cref{tab:app_thresholds}). A frozen threshold loses some BAcc relative to one tuned
on each source of the benchmark. For BARTSmiles LL the loss is 1.8 points on RP, 1.9 on ZR,
7.3 on LP, and 15.0 on GR.

\subsection{Statistical Testing}
\label{app:stats}

Unless otherwise specified, confidence intervals on a single metric are percentile
intervals from \num{10000} bootstrap resamples of the reactions in the evaluated subset.
The method-specific AUROCs in \cref{tab:results_per_method,tab:app_shallow_sep} instead
use DeLong intervals, as stated in their captions. AUROC differences between two systems
on the same reactions are tested with the paired DeLong
test~\citep{delong1988comparing}. \Cref{tab:app_delong} lists twelve paired DeLong tests,
with $p$-values adjusted by the Holm procedure~\citep{holm1979simple} over the twelve of
them together. One comparison survives correction: on RP, BARTSmiles LL exceeds \discname
by 11.1 AUROC points. The difference between \gbdtname and \discname on RP is
not significant; this does not establish equivalence.
Neither controlled pair of \cref{sec:negatives} reaches significance on any source,
including the 9.0-point difference on GR, where only \num{33} feasible candidates are
available. The per-method training comparison uses the separate paired-bootstrap
analysis and six-comparison correction specified in \cref{app:prereg}.

\begin{table}[h]
  \centering
  \caption{
    Paired DeLong tests, within source. $\Delta$ is AUROC(A) $-$ AUROC(B) in points.
    $p_{\text{adj}}$ is Holm-corrected over all twelve tests.
  }
  \label{tab:app_delong}
  \scriptsize
  \setlength{\tabcolsep}{4pt}
  \begin{tabular}{@{}l l l r l r r@{}}
    \toprule
    Source & A & B & $\Delta$ & 95\% CI & $p$ & $p_{\text{adj}}$ \\
    \midrule
    RP & BARTSmiles (LL) & \discname & $+11.1$ & $[+4.3, +17.9]$ & 0.0015 & \textbf{0.018} \\
    RP & \gbdtname & \discname & $+2.5$ & $[-4.3, +9.4]$ & 0.470 & 1.00 \\
    RP & BARTSmiles (LL) & \gbdtname & $+8.6$ & $[+0.8, +16.3]$ & 0.031 & 0.343 \\
    RP & \negrrabbr only & \discname & $+6.7$ & $[+0.3, +13.2]$ & 0.042 & 0.418 \\
    \addlinespace[2pt]
    RP & \negname & no negatives & $-0.7$ & $[-4.2, +2.7]$ & 0.671 & 1.00 \\
    ZR & \negname & no negatives & $+4.3$ & $[-2.5, +11.2]$ & 0.213 & 1.00 \\
    LP & \negname & no negatives & $+2.6$ & $[-3.2, +8.5]$ & 0.381 & 1.00 \\
    GR & \negname & no negatives & $+9.0$ & $[-3.0, +21.1]$ & 0.142 & 1.00 \\
    \addlinespace[2pt]
    RP & $+$ negatives, 3$\times$ & no negatives, 3$\times$ & $-4.0$ & $[-8.0, -0.0]$ & 0.050 & 0.445 \\
    ZR & $+$ negatives, 3$\times$ & no negatives, 3$\times$ & $+1.1$ & $[-3.3, +5.4]$ & 0.624 & 1.00 \\
    LP & $+$ negatives, 3$\times$ & no negatives, 3$\times$ & $+4.5$ & $[-2.0, +11.0]$ & 0.178 & 1.00 \\
    GR & $+$ negatives, 3$\times$ & no negatives, 3$\times$ & $+9.0$ & $[-3.4, +21.4]$ & 0.154 & 1.00 \\
    \bottomrule
  \end{tabular}
\end{table}

For the ranking of the eight retrosynthesis models in \cref{tab:app_ranking}, the interval
on Spearman's $\rho$ comes from \num{5000} bootstrap resamples of the proposals within each
retrosynthesis model, which propagates the uncertainty in both the expert accept rates and
the system's. The reference ordering is recomputed within each resample, so the
intervals account for sampling uncertainty in the expert ranking as well.

\input{tab/app_protocol}

%% file: tab/app_auxsens.tex
\begin{table}[h]
  \centering
  \caption{Source-mean scores after excluding the \num{104} feasible reactions
  annotated with \texttt{missing\_auxiliary\_reagent}, using the same frozen
  thresholds. Full AUROC is reproduced from \cref{tab:results_main};
  restricted AUROC and BAcc use the remaining reactions.}
  \label{tab:app_auxsens}
  \small
  \setlength{\tabcolsep}{5pt}
  \begin{tabular}{@{}l rrr@{}}
    \toprule
    System & Full AUROC & Restricted AUROC & Restricted BAcc \\
    \midrule
    Claude Opus 5        & --   & --   & 70.0 \\
    GPT-6 Astra          & --   & --   & 67.2 \\
    Chemformer (top-10)  & --   & --   & 64.5 \\
    Chemformer (LL)      & 72.6 & 74.6 & 65.7 \\
    BARTSmiles (top-10)  & --   & --   & 67.9 \\
    BARTSmiles (LL)      & 72.8 & 75.0 & 66.7 \\
    \negname             & 75.2 & 74.9 & 63.6 \\
    \discname            & 69.7 & 70.7 & 59.6 \\
    \gbdtname            & 70.7 & 74.7 & 66.6 \\
    \bottomrule
  \end{tabular}
\end{table}

%% file: tab/app_protocol.tex
\begin{table}[h]
  \centering
  \caption{
    Sensitivity of the discriminative verifier family to the checkpoint, in AUROC by source.
    \emph{Reported}: the fixed budget of \num{20000} steps used for this family.
    \emph{Final}: the last checkpoint of each run.
    \emph{Oracle}: the checkpoint with the highest average AUROC on \dataname, chosen with the
    benchmark labels. It is an optimistic upper bound over the evaluated checkpoints,
    rather than an independently selected performance estimate.
  }
  \label{tab:app_checkpoints}
  \scriptsize
  \setlength{\tabcolsep}{3.4pt}
  \begin{tabular}{@{}l l r ccccc@{}}
    \toprule
    Model & Rule & Step & RP & ZR & LP & GR & Avg. \\
    \midrule
    \negraabbr only & Reported & \num{20000} & 51.0 & 72.8 & 57.9 & 54.3 & 59.0 \\
     & Final & \num{22168} & 50.8 & 73.6 & 57.8 & 54.1 & 59.1 \\
     & Oracle & \num{11000} & 53.4 & 74.2 & 63.0 & 54.9 & 61.4 \\
    \addlinespace[2pt]
    \negrrabbr only & Reported & \num{20000} & 74.5 & 87.3 & 68.2 & 68.6 & 74.6 \\
     & Final & \num{22168} & 74.7 & 87.6 & 68.3 & 68.4 & 74.7 \\
     & Oracle & \num{5000} & 72.5 & 88.1 & 73.9 & 73.6 & 77.0 \\
    \addlinespace[2pt]
    \negncabbr only & Reported & \num{20000} & 54.1 & 68.9 & 60.5 & 58.3 & 60.4 \\
     & Final & \num{22168} & 53.8 & 69.3 & 60.7 & 58.4 & 60.6 \\
     & Oracle & \num{9000} & 55.4 & 65.0 & 62.7 & 60.7 & 61.0 \\
    \addlinespace[2pt]
    \negftabbr only & Reported & \num{20000} & 52.4 & 43.6 & 49.5 & 43.5 & 47.2 \\
     & Final & \num{22168} & 52.3 & 42.9 & 49.8 & 43.4 & 47.1 \\
     & Oracle & \num{6000} & 53.2 & 49.9 & 53.5 & 48.5 & 51.3 \\
    \addlinespace[2pt]
    \negetabbr only & Reported & \num{20000} & 52.0 & 50.3 & 58.1 & 58.0 & 54.6 \\
     & Final & \num{22168} & 52.0 & 49.1 & 58.1 & 57.7 & 54.2 \\
     & Oracle & \num{4000} & 59.8 & 63.6 & 61.2 & 60.4 & 61.2 \\
    \addlinespace[2pt]
    Mixture, 1.42M pairs & Reported & \num{20000} & 68.0 & 75.7 & 66.8 & 61.2 & 67.9 \\
     & Final & \num{22168} & 67.8 & 76.0 & 67.1 & 60.3 & 67.8 \\
     & Oracle & \num{7000} & 65.6 & 90.3 & 64.5 & 67.1 & 71.9 \\
    \addlinespace[2pt]
    Mixture, 7.10M pairs & Reported & \num{20000} & 67.8 & 79.0 & 66.0 & 66.0 & 69.7 \\
     & Final & \num{41571} & 66.9 & 74.0 & 64.6 & 61.0 & 66.6 \\
     & Oracle & \num{10000} & 66.8 & 79.4 & 65.5 & 73.2 & 71.2 \\
    \bottomrule
  \end{tabular}
\end{table}

\begin{table}[h]
  \centering
  \caption{
    Frozen decision thresholds. Each maximizes balanced accuracy on the \num{10000} held-out
    generated pairs and is applied unchanged to \dataname. Scores are log-likelihoods for the
    forward models, raw logits for the verifier models, and probabilities for the shallow
    baseline.
  }
  \label{tab:app_thresholds}
  \scriptsize
  \begin{tabular}{@{}l r@{}}
    \toprule
    System & Threshold \\
    \midrule
    \gbdtname & $0.5101$ \\
    Chemformer (LL) & $-0.0787$ \\
    BARTSmiles (LL) & $-0.0829$ \\
    \discname & $-0.0913$ \\
    BARTSmiles (LL), extra epoch & $-0.0942$ \\
    BARTSmiles (LL $+$ neg.) & $-0.4863$ \\
    BARTSmiles (LL), extra epoch, 3$\times$ & $-0.1354$ \\
    BARTSmiles (LL $+$ neg.), 3$\times$ & $-0.5464$ \\
    Mixture, 1.42M pairs & $-0.0684$ \\
    \negraabbr only & $3.3424$ \\
    \negrrabbr only & $2.1345$ \\
    \negncabbr only & $5.4880$ \\
    \negftabbr only & $3.8126$ \\
    \negetabbr only & $3.7918$ \\
    \bottomrule
  \end{tabular}
\end{table}

%% file: appendices/additional_results.tex
\section{Additional Results}
\label{app:full_results}

\paragraph{Full results by source.}
\Cref{tab:app_full_auroc} gives AUROC with bootstrap confidence intervals and BAcc for every
system, including the rule-based baseline and both controlled pairs, and \cref{tab:app_rates}
gives the TPR and TNR behind each BAcc. The intervals on ZR and LP are wide because these
sources contain only 14 and 32 examples of their minority label.
For comparison with evaluation on constructed labels, \discname reaches
98.8 AUROC on held-out generated pairs but 69.7 mean AUROC on \dataname.

\paragraph{Generated reactions by method, at each operating point.}
\Cref{tab:app_per_method_rates} gives the share of each method's infeasible candidates that every
system rejects, and the share of the \num{33} feasible GR candidates it accepts,
complementing the threshold-free comparison of \cref{tab:results_per_method}. Systems
calibrated on generated pairs accept few feasible GR candidates. On the
calibration data, generated candidates are assigned negative labels by construction.

\paragraph{LLM perturbations by producing model.}
LP was constructed with GPT-5.4 and Claude Opus 4.6, and we evaluate newer models of the
same two families. \Cref{tab:app_lp_generator} splits LP by the model that produced each
perturbation. Neither evaluated LLM has higher BAcc on the perturbations of
its own family in this sample.

\paragraph{The verifier family by method.}
\Cref{tab:app_family_matrix} gives the AUROC of the discriminative verifiers
of \cref{app:family} and the controlled pairs on the candidates of each method.
Unlike \cref{tab:results_per_method}, each column here uses only the feasible
candidates from that same method, rather than the shared set of 33 references.
The two tables therefore answer different comparison questions.
A verifier is often not the best on its own method. For example, the verifier trained on
\negrrabbr candidates scores higher on \negraabbr candidates than the one trained on them.

\input{tab/app_results}

\subsection{Ranking Retrosynthesis Models}
\label{app:ranking}
\label{subsec:ranking}

\input{tab/app_ranking}

Beyond screening individual reactions, verifiers can be used to compare
retrosynthesis models. We therefore examine whether their acceptance rates
yield the same model ordering as expert labels. For each of the eight models
represented in the 282-reaction RP subset, we calculate verifier and expert
acceptance rates, then compare the resulting rankings using Spearman's
correlation (\cref{tab:app_ranking}). Rank correlations cover the LLM
decisions and thresholded continuous scores; the top-10 rows report
acceptance rate and precision only.

\textbf{The LLM-based rankings agree most closely with the expert ordering
among the evaluated rankings.}
Experts accept 48.3--92.6\% of each model's sampled proposals.
Spearman's $\rho$ is 0.93 for Claude (95\% CI $[0.31,0.96]$) and 0.76
for GPT ($[0.26,0.93]$). These intervals incorporate uncertainty in the expert
ordering (\cref{app:stats}). The thresholded
continuous-score systems have correlations between $-0.19$ and $+0.17$,
with intervals spanning zero. They include BARTSmiles LL ($\rho = +0.01$),
whose RP BAcc of 71.5 is close to Claude's 69.9. Thus, strong reaction-level
RP scores do not necessarily imply agreement with expert model rankings
at the thresholds used here.

This analysis covers eight models and their sampled proposals at the
evaluated operating points. The wide intervals limit conclusions about
the relative quality of verifiers as judges; these results do not establish
a general ranking for that use.

\paragraph{Filtering tradeoff.}
The same analysis quantifies the tradeoff when these decisions are used
to filter proposals. BARTSmiles LL retains 50.0\% of RP proposals, of which
87.9\% are feasible. Claude retains 70.2\%, of which 81.8\% are feasible.
Precision must therefore be considered together with the fraction retained
when evaluating a verifier for planning.

%% file: tab/app_results.tex
\begin{table}[h]
  \centering
  \caption{
    AUROC with 95\% bootstrap confidence intervals (\num{10000} resamples) and BAcc by source
    for all systems, including the two controlled pairs of \cref{sec:negatives}.
  }
  \label{tab:app_full_auroc}
  \scriptsize
  \setlength{\tabcolsep}{2.6pt}
  \begin{tabular}{@{}l cc cc cc cc@{}}
    \toprule
    & \multicolumn{2}{c}{RP} & \multicolumn{2}{c}{ZR} & \multicolumn{2}{c}{LP} & \multicolumn{2}{c}{GR} \\
    \cmidrule(lr){2-3}\cmidrule(lr){4-5}\cmidrule(lr){6-7}\cmidrule(l){8-9}
    System & AUROC & BAcc & AUROC & BAcc & AUROC & BAcc & AUROC & BAcc \\
    \midrule
    Claude Opus 5 & -- & 69.9 & -- & 64.3 & -- & 76.1 & -- & 61.8 \\
    GPT-6 Astra & -- & 66.8 & -- & 63.7 & -- & 69.6 & -- & 62.7 \\
    Chemformer (top-10) & -- & 70.6 & -- & 67.2 & -- & 63.3 & -- & 50.1 \\
    Chemformer (LL) & 76.7 {\scriptsize\color{gray}[70.6, 82.2]} & 71.6 & 82.4 {\scriptsize\color{gray}[69.9, 93.0]} & 70.2 & 64.5 {\scriptsize\color{gray}[50.4, 77.6]} & 63.8 & 67.0 {\scriptsize\color{gray}[58.8, 74.6]} & 49.8 \\
    BARTSmiles (top-10) & -- & 73.1 & -- & 74.4 & -- & 59.0 & -- & 57.1 \\
    BARTSmiles (LL) & 78.9 {\scriptsize\color{gray}[73.1, 84.2]} & 71.5 & 79.5 {\scriptsize\color{gray}[64.8, 92.1]} & 72.5 & 66.1 {\scriptsize\color{gray}[52.8, 78.6]} & 60.4 & 66.9 {\scriptsize\color{gray}[57.5, 75.8]} & 53.4 \\
    \aizname & 57.0 {\scriptsize\color{gray}[49.6, 64.3]} & 56.6 & 74.0 {\scriptsize\color{gray}[55.7, 89.5]} & 69.9 & 62.3 {\scriptsize\color{gray}[49.6, 74.3]} & 57.1 & 43.1 {\scriptsize\color{gray}[33.9, 52.8]} & 42.6 \\
    \negname & 77.0 {\scriptsize\color{gray}[71.1, 82.4]} & 67.1 & 80.8 {\scriptsize\color{gray}[67.9, 92.0]} & 63.7 & 69.4 {\scriptsize\color{gray}[56.8, 80.9]} & 66.1 & 73.5 {\scriptsize\color{gray}[64.2, 82.4]} & 54.9 \\
    \discname & 67.8 {\scriptsize\color{gray}[61.0, 74.4]} & 62.5 & 79.0 {\scriptsize\color{gray}[63.1, 92.1]} & 60.1 & 66.0 {\scriptsize\color{gray}[53.2, 78.4]} & 54.8 & 66.0 {\scriptsize\color{gray}[55.1, 76.3]} & 56.6 \\
    \gbdtname & 70.3 {\scriptsize\color{gray}[63.8, 76.4]} & 64.0 & 72.2 {\scriptsize\color{gray}[59.4, 83.6]} & 56.3 & 75.4 {\scriptsize\color{gray}[64.1, 85.3]} & 70.2 & 64.7 {\scriptsize\color{gray}[53.9, 75.4]} & 62.4 \\
    \rulemethodname & -- & 60.0 & -- & 61.5 & -- & 58.0 & -- & 61.3 \\
    BARTSmiles (LL), extra epoch & 77.8 {\scriptsize\color{gray}[71.9, 83.3]} & 71.5 & 76.5 {\scriptsize\color{gray}[59.9, 90.7]} & 70.2 & 66.8 {\scriptsize\color{gray}[53.6, 79.3]} & 58.9 & 64.5 {\scriptsize\color{gray}[55.3, 73.2]} & 49.1 \\
    BARTSmiles (LL), extra epoch, 3$\times$ & 78.2 {\scriptsize\color{gray}[72.5, 83.5]} & 71.4 & 78.5 {\scriptsize\color{gray}[63.1, 91.6]} & 74.4 & 65.3 {\scriptsize\color{gray}[52.1, 77.8]} & 62.6 & 64.5 {\scriptsize\color{gray}[55.3, 73.1]} & 50.7 \\
    BARTSmiles (LL $+$ neg.), 3$\times$ & 74.2 {\scriptsize\color{gray}[68.2, 80.0]} & 65.5 & 79.6 {\scriptsize\color{gray}[65.6, 91.4]} & 64.3 & 69.8 {\scriptsize\color{gray}[57.4, 80.9]} & 65.8 & 73.5 {\scriptsize\color{gray}[64.3, 82.2]} & 56.6 \\
    \bottomrule
  \end{tabular}
\end{table}

\begin{table}[h]
  \centering
  \caption{
    TPR and TNR by source at each system's operating point. TPR is the share of feasible
    reactions accepted, TNR the share of infeasible reactions rejected.
  }
  \label{tab:app_rates}
  \scriptsize
  \setlength{\tabcolsep}{3.2pt}
  \begin{tabular}{@{}l cc cc cc cc@{}}
    \toprule
    & \multicolumn{2}{c}{RP (197/85)} & \multicolumn{2}{c}{ZR (79/14)} & \multicolumn{2}{c}{LP (32/55)} & \multicolumn{2}{c}{GR (33/256)} \\
    \cmidrule(lr){2-3}\cmidrule(lr){4-5}\cmidrule(lr){6-7}\cmidrule(l){8-9}
    System & TPR & TNR & TPR & TNR & TPR & TNR & TPR & TNR \\
    \midrule
    Claude Opus 5 & 82.2 & 57.6 & 100.0 & 28.6 & 61.3 & 90.9 & 33.3 & 90.2 \\
    GPT-6 Astra & 88.8 & 44.7 & 98.7 & 28.6 & 71.9 & 67.3 & 51.5 & 73.8 \\
    Chemformer (top-10) & 77.7 & 63.5 & 98.7 & 35.7 & 37.5 & 89.1 & 18.2 & 82.0 \\
    Chemformer (LL) & 64.5 & 78.8 & 97.5 & 42.9 & 31.2 & 96.4 & 12.1 & 87.5 \\
    BARTSmiles (top-10) & 79.2 & 67.1 & 98.7 & 50.0 & 34.4 & 83.6 & 33.3 & 80.9 \\
    BARTSmiles (LL) & 62.9 & 80.0 & 94.9 & 50.0 & 28.1 & 92.7 & 18.2 & 88.7 \\
    \aizname & 57.9 & 55.3 & 68.4 & 71.4 & 46.9 & 67.3 & 12.1 & 73.0 \\
    \negname & 73.1 & 61.2 & 98.7 & 28.6 & 59.4 & 72.7 & 12.1 & 97.7 \\
    \discname & 75.6 & 49.4 & 98.7 & 21.4 & 75.0 & 34.5 & 18.2 & 94.9 \\
    \gbdtname & 68.0 & 60.0 & 91.1 & 21.4 & 75.0 & 65.5 & 42.4 & 82.4 \\
    \rulemethodname & 88.3 & 31.8 & 87.3 & 35.7 & 68.8 & 47.3 & 90.9 & 31.6 \\
    BARTSmiles (LL), extra epoch & 62.9 & 80.0 & 97.5 & 42.9 & 25.0 & 92.7 & 9.1 & 89.1 \\
    BARTSmiles (LL), extra epoch, 3$\times$ & 67.5 & 75.3 & 98.7 & 50.0 & 34.4 & 90.9 & 15.2 & 86.3 \\
    BARTSmiles (LL $+$ neg.), 3$\times$ & 72.1 & 58.8 & 100.0 & 28.6 & 62.5 & 69.1 & 15.2 & 98.0 \\
    \bottomrule
  \end{tabular}
\end{table}

\begin{table}[h]
  \centering
  \caption{
    Behavior on GR at each system's operating point. The first five columns give the
    share of each method's infeasible candidates that a system rejects (counts in
    parentheses), the last the share of the \num{33} feasible GR candidates it accepts.
    Both depend on where the threshold falls, and \cref{tab:results_per_method} gives
    the threshold-free comparison for the systems that return a score.
  }
  \label{tab:app_per_method_rates}
  \scriptsize
  \setlength{\tabcolsep}{4pt}
  \begin{tabular}{@{}l ccc cc c@{}}
    \toprule
    & \multicolumn{3}{c}{Reactants changed} & \multicolumn{2}{c}{Products changed} & Feasible \\
    \cmidrule(lr){2-4}\cmidrule(lr){5-6}\cmidrule(l){7-7}
    System & \negraabbr{\scriptsize(50)} & \negrrabbr{\scriptsize(53)} & \negncabbr{\scriptsize(54)} & \negftabbr{\scriptsize(44)} & \negetabbr{\scriptsize(55)} & {\scriptsize(33)} \\
    \midrule
    \multicolumn{7}{@{}l}{\textit{Large language models}} \\
    Claude Opus 5 & 84.0 & 98.1 & 83.3 & 90.9 & 94.5 & 33.3 \\
    GPT-6 Astra & 54.0 & 96.2 & 40.7 & 84.1 & 94.5 & 51.5 \\
    \midrule
    \multicolumn{7}{@{}l}{\textit{Forward-model proxies}} \\
    Chemformer (top-10) & 70.0 & 98.1 & 44.4 & 100.0 & 100.0 & 18.2 \\
    Chemformer (LL) & 82.0 & 100.0 & 57.4 & 100.0 & 100.0 & 12.1 \\
    BARTSmiles (top-10) & 66.0 & 98.1 & 42.6 & 100.0 & 100.0 & 33.3 \\
    BARTSmiles (LL) & 84.0 & 100.0 & 61.1 & 100.0 & 100.0 & 18.2 \\
    \midrule
    \multicolumn{7}{@{}l}{\textit{Deployed filter}} \\
    \aizname & 74.0 & 64.2 & 46.3 & 95.5 & 89.1 & 12.1 \\
    \midrule
    \multicolumn{7}{@{}l}{\textit{Trained on our generated negative candidates}} \\
    \negname & 96.0 & 94.3 & 100.0 & 97.7 & 100.0 & 12.1 \\
    \discname & 94.0 & 90.6 & 100.0 & 93.2 & 96.4 & 18.2 \\
    \gbdtname & 76.0 & 56.6 & 100.0 & 84.1 & 94.5 & 42.4 \\
    \bottomrule
  \end{tabular}
\end{table}

\begin{table}[h]
  \centering
  \caption{
    Results on LP split by the LLM that produced the perturbation (feasible/infeasible
    counts in parentheses). The evaluated LLMs are newer models of the same two families.
    Neither shows an advantage on perturbations produced by its own family. One Claude Opus 5
    response on a GPT-5.4 perturbation could not be parsed and is excluded.
  }
  \label{tab:app_lp_generator}
  \scriptsize
  \setlength{\tabcolsep}{4pt}
  \begin{tabular}{@{}l ccc ccc@{}}
    \toprule
    & \multicolumn{3}{c}{Produced by GPT-5.4 (15/30)} & \multicolumn{3}{c}{Produced by Claude Opus 4.6 (17/25)} \\
    \cmidrule(lr){2-4}\cmidrule(l){5-7}
    System & BAcc & TPR & TNR & BAcc & TPR & TNR \\
    \midrule
    Claude Opus 5 & 76.9 & 57.1 & 96.7 & 74.4 & 64.7 & 84.0 \\
    GPT-6 Astra & 63.3 & 46.7 & 80.0 & 73.1 & 94.1 & 52.0 \\
    Chemformer (top-10) & 65.0 & 46.7 & 83.3 & 62.7 & 29.4 & 96.0 \\
    Chemformer (LL) & 66.7 & 40.0 & 93.3 & 61.8 & 23.5 & 100.0 \\
    BARTSmiles (top-10) & 53.3 & 33.3 & 73.3 & 65.6 & 35.3 & 96.0 \\
    BARTSmiles (LL) & 61.7 & 33.3 & 90.0 & 59.8 & 23.5 & 96.0 \\
    \aizname & 55.0 & 46.7 & 63.3 & 59.5 & 47.1 & 72.0 \\
    \negname & 66.7 & 60.0 & 73.3 & 65.4 & 58.8 & 72.0 \\
    \discname & 58.3 & 80.0 & 36.7 & 51.3 & 70.6 & 32.0 \\
    \gbdtname & 68.3 & 80.0 & 56.7 & 73.3 & 70.6 & 76.0 \\
    \bottomrule
  \end{tabular}
\end{table}

\begin{table}[h]
  \centering
  \caption{
    AUROC on GR by negative candidate generation method for the discriminative
    verifier family of \cref{app:family} and the two controlled pairs of
    \cref{app:controlled_pair}. Each cell compares one method's infeasible candidates with the
    feasible candidates of the same method (4--13 of them), so single cells are imprecise.
    Italic cells are a single-method model evaluated on its own method.
  }
  \label{tab:app_family_matrix}
  \scriptsize
  \setlength{\tabcolsep}{4.5pt}
  \begin{tabular}{@{}l ccccc@{}}
    \toprule
    Trained on & \negraabbr & \negrrabbr & \negncabbr & \negftabbr & \negetabbr \\
    \midrule
    \negraabbr only & \textit{65.5} & 55.7 & 81.5 & 37.1 & 71.4 \\
    \negrrabbr only & 72.5 & \textit{72.2} & 30.6 & 73.8 & 80.9 \\
    \negncabbr only & 52.5 & 51.9 & \textit{81.0} & 61.7 & 42.7 \\
    \negftabbr only & 61.0 & 49.5 & 33.3 & \textit{45.8} & 72.7 \\
    \negetabbr only & 43.0 & 74.1 & 40.7 & 64.9 & \textit{55.0} \\
    Mixture, 1.42M pairs & 52.0 & 93.4 & 87.5 & 54.4 & 83.6 \\
    Mixture, 7.10M pairs & 45.2 & 84.0 & 90.3 & 65.0 & 80.9 \\
    \midrule
    BARTSmiles (LL), extra epoch & 66.5 & 57.1 & 25.5 & 84.3 & 82.3 \\
    BARTSmiles (LL $+$ neg.) & 51.7 & 75.0 & 92.1 & 83.0 & 77.3 \\
    BARTSmiles (LL), extra epoch, 3$\times$ & 68.0 & 56.6 & 29.6 & 83.7 & 80.9 \\
    BARTSmiles (LL $+$ neg.), 3$\times$ & 54.2 & 78.8 & 94.4 & 82.0 & 77.7 \\
    \bottomrule
  \end{tabular}
\end{table}

%% file: tab/app_ranking.tex
\begin{table}[h]
  \centering
  \caption{Ranking the eight retrosynthesis models by the share of their RP proposals a
  system accepts. \emph{Accept} is that share over all \num{282} proposals, against the
  \num{69.9}\% the experts accept, and \emph{Precision} the share of accepted proposals
  that are feasible. $\rho$ is the Spearman correlation between the system's ranking of
  the eight models and the experts' ranking; brackets give 95\% intervals from
  \num{5000} bootstrap resamples of the proposals within each model.
  Dashes denote unreported top-10 rank correlations.}
  \label{tab:app_ranking}
  \small
  \setlength{\tabcolsep}{5pt}
  \begin{tabular}{@{}l rr l@{}}
    \toprule
    System & Accept (\%) & Precision (\%) & $\rho$ vs.\ experts \\
    \midrule
    \textit{Expert labels} & \textit{69.9} & \textit{100.0} & --- \\
    \midrule
    Claude Opus 5 & 70.2 & 81.8 & $+0.93$ {\scriptsize\color{gray}[$+0.31$, $+0.96$]} \\
    GPT-6 Astra & 78.7 & 78.8 & $+0.76$ {\scriptsize\color{gray}[$+0.26$, $+0.93$]} \\
    Chemformer (top-10) & 65.2 & 83.2 & --- \\
    Chemformer (LL) & 51.4 & 87.6 & $-0.04$ {\scriptsize\color{gray}[$-0.68$, $+0.69$]} \\
    BARTSmiles (top-10) & 65.2 & 84.8 & --- \\
    BARTSmiles (LL) & 50.0 & 87.9 & $+0.01$ {\scriptsize\color{gray}[$-0.67$, $+0.67$]} \\
    \negname & 62.8 & 81.4 & $-0.19$ {\scriptsize\color{gray}[$-0.67$, $+0.52$]} \\
    \discname & 68.1 & 77.6 & $-0.12$ {\scriptsize\color{gray}[$-0.47$, $+0.66$]} \\
    \gbdtname & 59.6 & 79.8 & $-0.07$ {\scriptsize\color{gray}[$-0.54$, $+0.69$]} \\
    \aizname & 53.9 & 75.0 & $+0.17$ {\scriptsize\color{gray}[$-0.61$, $+0.64$]} \\
    \bottomrule
  \end{tabular}
\end{table}

%% file: appendices/data_card.tex
\section{Data Card}
\label{app:datacard}

\paragraph{Contents.}
The release has three parts. The \textbf{benchmark} is the \goldN{} expert-labeled
reactions of \cref{sec:benchmark}. Each record holds the canonical reactant and product
SMILES, the binary feasibility label, the source and the model or method that produced the
reaction, a failure reason for infeasible reactions where one was recorded
(\cref{app:reasons}), and free-text notes where the annotators left them. The
\textbf{generated corpus} contains negative candidates paired with their parent
reactions, each carrying the method that produced it, split into training and
test portions that share no reaction (\cref{app:splits,app:corpus_stats}).
The \textbf{protocol} is the
annotation guideline given to the chemists, the decision procedure of
\cref{app:procedure}, and the failure-reason taxonomy.

\paragraph{Collection and labeling.}
Reactions derive from ORDerly~\citep{wigh2024orderly}, a curated subset of the Open
Reaction Database~\citep{kearnes2021ord}, which draws on patents and published literature.
No new experiments were run and no human subjects are involved. Two chemists among the
authors labeled reactions without seeing their sources, following the calibration,
independent annotation and adjudication procedure in \cref{app:annotation}.

\paragraph{Intended use.}
The benchmark supports evaluation of reaction feasibility verifiers under
\cref{sec:task}. The generated corpus supports training and comparisons of
negative generation methods, as studied in \cref{sec:negatives}.
Annotation quality is documented in \cref{app:agreement}; coverage and
application scope are described in \cref{app:limitations}.

\paragraph{Distribution and maintenance.}
The benchmark is available at
\url{https://huggingface.co/datasets/FREA-benchmark/FREA-benchmark},
and the generated corpus at
\url{https://huggingface.co/datasets/FREA-benchmark/FREA-corpus}.
The benchmark and the generated corpus are released under CC BY 4.0. A changelog will record label corrections,
and we expect the benchmark to grow as annotation continues.

%% file: appendices/limitations.tex
\section{Limitations}
\label{app:limitations}
\label{sec:limitations}

\paragraph{Evaluation scope.}
\dataname evaluates whether individual proposals admit reasonable conditions
under the stated completeness and reactivity criterion, without specifying
those conditions or testing the reactions experimentally.
A feasible label therefore does not establish a practical procedure, an
acceptable yield, or the quality of a complete synthesis route, where
compatibility between steps and purification also matter.

\paragraph{Expert judgment and annotation boundaries.}
Even with a written criterion, chemists may differ on whether conditions
are reasonable, a target is isolable, or an omitted component serves a
permitted auxiliary role in a particular reaction.
Our labels rely on two chemists, whose independent agreement was
\goldAgreePct{} with Cohen's $\kappa = \goldKappa$ before adjudication.
Joint review produces a shared reference, but does not eliminate possible
annotation errors or establish agreement with other experts; the final
binary labels also do not express the uncertainty of borderline judgments.

\paragraph{Coverage and source composition.}
All four sources derive from ORDerly's patent and literature records, and
screening removes parsing failures and selected structural defects before
annotation, so the benchmark does not represent the full range of chemistry
or unfiltered planner outputs.
The source proportions are chosen for evaluation, rather than estimated
from deployment, and the four-source mean need not reflect performance in
a particular planning workflow.
GR also uses the same generation methods as our negative training corpus,
so gains on this source do not by themselves demonstrate transfer to model
proposals or experimental records.

\paragraph{Sample size and comparison design.}
Several comparisons depend on small minority classes: ZR has only 14
infeasible reactions, and the per-method GR analyses share 33 feasible
reference reactions.
Those references come from all five generation methods, so each per-method AUROC
measures discrimination against a mixed reference set and can reflect
differences between generation methods as well as feasibility.
The reported confidence intervals condition on the fitted models and final
labels; they do not account for annotation errors or variation across
training runs.

\paragraph{Generated labels and exploratory training.}
Generated candidates can remain feasible, and treating them as negatives
during training can penalize valid alternatives; differences in candidate
purity also prevent us from attributing performance differences solely to
the generation method.
Our controlled training comparisons use one backbone, one corpus and one
seed per setting, and most observed gains are not statistically significant
after correction for multiple comparisons.
The resulting hypotheses about label reliability and transfer therefore
require tests across seeds, architectures and corpora.

\paragraph{Decision rules and model comparisons.}
Thresholds calibrated on generated pairs may transfer poorly to other
sources, so a low acceptance rate at these thresholds does not establish
that a model cannot achieve a different precision--recall tradeoff.
The LLM results likewise describe two models under one shared prompt, and
may change with the prompt or inference settings.
The retrosynthesis ranking analysis covers eight models on their sampled
proposals at the evaluated operating points, rather than establishing a
general ordering of planners or verifiers.